\documentclass{article} 
\usepackage{iclr2027_conference,times}

\usepackage{amsmath,amsfonts,bm}

\def\eqref#1{equation~\ref{#1}}

\def\1{\bm{1}}

\DeclareMathAlphabet{\mathsfit}{\encodingdefault}{\sfdefault}{m}{sl}
\SetMathAlphabet{\mathsfit}{bold}{\encodingdefault}{\sfdefault}{bx}{n}

\usepackage{hyperref}
\usepackage{url}
\usepackage{graphicx}
\usepackage{subcaption}
\usepackage{booktabs}
\usepackage[table]{xcolor}
\usepackage{multirow}
\usepackage{wrapfig}
\newcommand{\cmark}{\ensuremath{\surd}}
\newcommand{\xmark}{\ensuremath{\times}}

\definecolor{oursbg}{HTML}{EAF6F4}
\title{MarvisNav: Making Memory Visible on Route Choices for Zero-Shot Object Navigation}

\author{%
\textbf{Jincheng Wang\textsuperscript{1}, Chi Pui Chan\textsuperscript{1}, Wei Zeng\textsuperscript{2}}\\
\textbf{Shuyang Zhang\textsuperscript{3}, Jianhao Jiao\textsuperscript{2}\textsuperscript{*}, Dimitrios Kanoulas\textsuperscript{1}}\\[0.4em]
{\normalfont \textsuperscript{1}University College London, University of London}\\
{\normalfont \textsuperscript{2}China Merchants Group, LionRock AI Lab}\\
{\normalfont \textsuperscript{3}Shenzhen University}\\
{\normalfont \textsuperscript{*}Corresponding Author}
}
\hypersetup{
  pdftitle={MarvisNav: Making Memory Visible on Route Choices for Zero-Shot Object Navigation},
  pdfauthor={Jincheng Wang, Chi Pui Chan, Wei Zeng, Shuyang Zhang, Jianhao Jiao, Dimitrios Kanoulas}
}

\iclrfinalcopy 
\begin{document}

\maketitle
\lhead{Preprint}
\vspace{-0.49cm}
\begin{abstract}

When searching for an object, people choose their next move by considering both likely target locations and places already explored. The current view can cue place-associated memories, bringing target relevance and prior exploration into the same spatial context. In many zero-shot object navigation (ZSON) methods, however, vision-language models (VLMs) infer promising search areas from egocentric images, but exploration history enters the decision through separate representations, \textit{e.g.,} as text or maps.
This separation either requires an additional fusion step to align semantic evidence with exploration state, or leaves the correspondence between memory and route choices implicit for the VLM to recover.
We instead make exploration memory visible directly on visual route choice.
We propose MarvisNav, a ZSON framework that maintains a topological graph and projects candidate nodes together with their exploration states onto the egocentric perception as memory-bearing visual route choices.
These topological states capture local exploration progress beyond binary visitation. By binding exploration state directly to each visual candidate, MarvisNav enables the VLM to jointly evaluate target relevance and exploration state and make the final high-level selection without a separate post-hoc fusion or reranking stage. Without policy training, MarvisNav achieves state-of-the-art performance on HM3D ($81.2\%$ SR and $42.5\%$ SPL), while competitive on MP3D. Moreover, it outperforms representative VLM-based methods with far fewer VLM calls (\textit{e.g.,} $7.5\%$ of WMNav). Real-robot experiments across diverse scenes further validate its practical deployability. Beyond MarvisNav, our study shows that memory representation shapes VLM decisions and ZSON performance, highlighting that effective memory use in ZSON depends not only on its availability, but also on how it is represented.
Code will be released at https://wangjincheng1998.github.io/MarvisNav/

\end{abstract}


\section{Introduction}
Imagine a robot asked to find an office chair in a home it has never seen. To succeed, it needs to infer where the target is likely to be and explore efficiently. This challenge is formalized as Object Navigation. Standard supervised methods are typically trained on a closed set of target categories~\citep{chaplot2020object,chaplot2021seal,poni,HIEM}, limiting their ability to generalize to new objects. Zero-shot Object Navigation (ZSON) instead leverages pretrained semantic knowledge to search for novel targets without category-specific training~\citep{zson,chen2023not,CoWs,openfmnav}, making it a key step toward open-world tasks.

ZSON is a long-horizon search problem under partial observability~\citep{2025cvpr}. When the target is not visible, the agent needs to use semantic cues to infer promising search areas. Existing methods obtain such guidance through visual-semantic alignment, which scores target relevance from visual observations
~\citep{CoWs,vlfm,apexnav}, or contextual semantic reasoning, which exploits scene context and spatial relations~\citep{esc,sgnav,cognav}. Recently, generative Vision-Language Models (VLMs) have been widely adopted for object search, leveraging strong multimodal understanding and broad knowledge for richer semantic and spatial reasoning over observations. Long-horizon search, however, requires a sequence of high-level decisions. A VLM does not by itself maintain the agent's exploration state across successive decisions. VLM-based methods therefore maintain explicit memory~\citep{wmnav,openfrontier} to avoid re-selecting semantically promising but already searched areas.



\begin{figure}[t]
    \centering
    \includegraphics[width=0.97\linewidth]{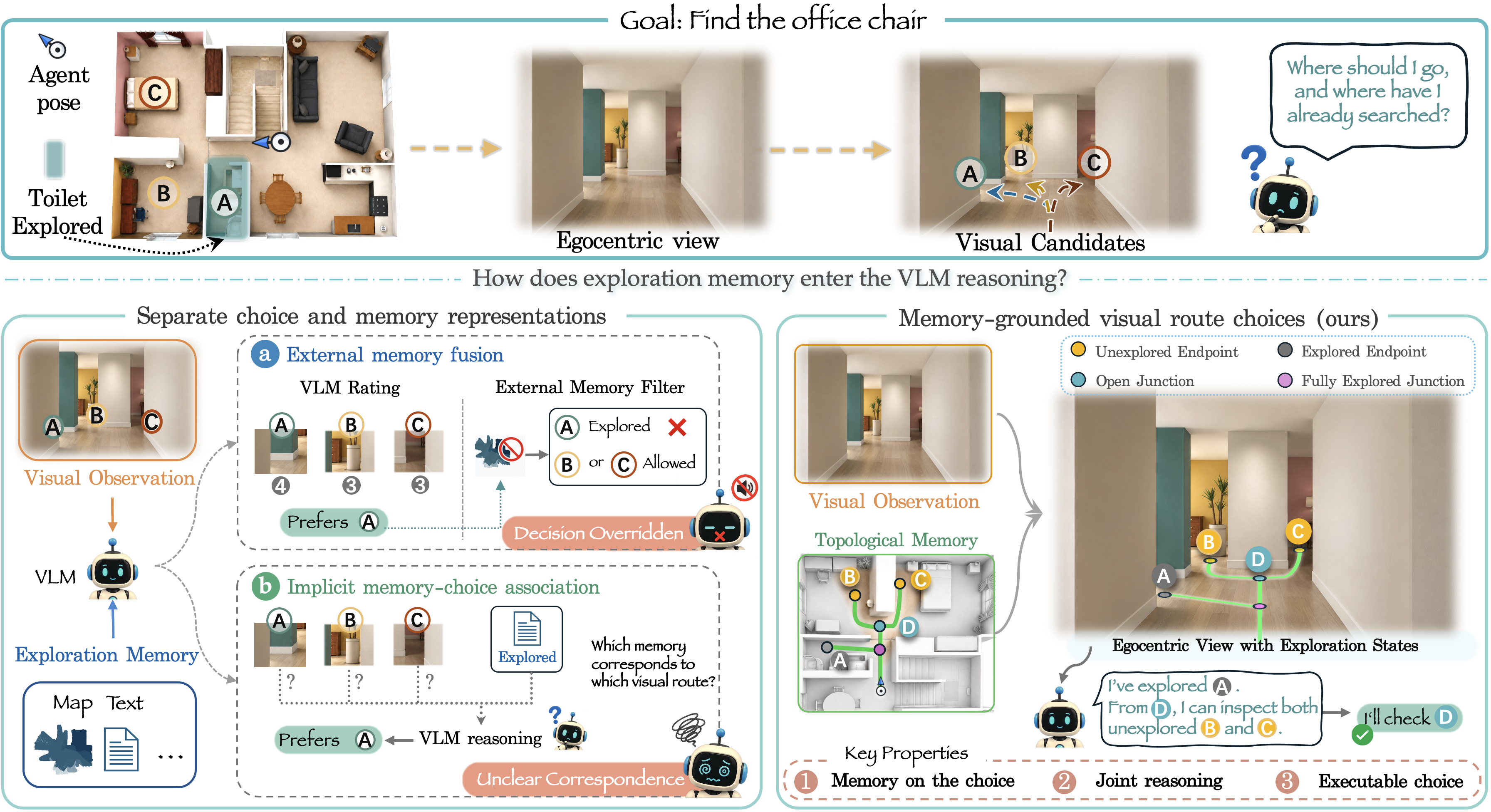}
    \caption{\textbf{Top:} Object search requires a VLM to consider both where the target might be found and which areas have already been searched. \textbf{Left:} With separate representations, memory either modifies the VLM's preference externally or leaves the VLM to infer its correspondence with the choices implicitly. \textbf{Right:} We instead ground topological exploration memory onto egocentric route choices, enabling joint reasoning over target relevance and exploration state. Junctions such as \textbf{D} are also valid choices. The selected choice directly specifies a navigation goal.}
    \label{fig1}
\vspace{-0.1cm}
\end{figure}

Semantic guidance and exploration memory, however, are often met in
separate representations. VLMs form semantic judgments from the current egocentric view, whereas exploration memory may be carried as separate textual histories or dense maps. Bringing the two together therefore requires an additional fusion or alignment step before decision-making. When memory remains outside the VLM, the system must explicitly define how it changes the model's semantic judgment~\citep{wmnav,openfrontier}, for example through hand-crafted candidate filtering or score adjustment. This places part of the choice outside the VLM and may constrain or override its judgment. When memory is instead supplied to the VLM, external fusion may be avoided, but the alignment problem remains. A textual history records past events in language, whereas a top-down map lays them out in an allocentric frame~\citep{wu2026rememnav,History-Augmented,habibpour2025think}. Neither is expressed in the egocentric view the agent is reasoning over. The VLM must therefore infer how remembered places relate to the current view before using that memory to decide where to search next (Fig.~\ref{fig1}, left).

Consider how a person searches for the chair in an unfamiliar house. They make both judgments from the same view, seeing which doorways look promising and which lead to areas already searched. A searched room is dismissed at a glance, while a doorway seen earlier remains a candidate if it still opens onto unexplored space. By making exploration cues visible on the corresponding spatial choices in views, we may then be able to give VLMs a similar decision interface, so that semantic evidence and exploration state are read together rather than aligned across representations.


To this end, we introduce \textbf{MarvisNav}, a ZSON framework that \textbf{ma}kes exploration memo\textbf{r}y \textbf{vis}ible to VLMs by grounding it directly on visual route choices.  We first reduce continuous exploration into sparse high-level decisions, where the VLM selects among navigable search directions. We derive these choices from a Reduced Voronoi Graph (RVG)~\citep{voronav}, where selected junctions serve as decision sites and local paths define candidate directions associated with reachable nodes. We extend this representation into a State-Aware RVG that associates each candidate with a local exploration state indicating whether nearby connected space may remain unexplored. At each decision, MarvisNav projects visible RVG paths into the egocentric RGB views and annotates candidates with local identifiers and exploration states, forming memory-bearing visual route choices (Fig.~\ref{fig1}, right). A VLM then jointly evaluates target relevance and exploration state on these choices and selects an identifier in a single query. The selected identifier directly specifies the next navigation goal, enabling memory-conditioned route selection without additional semantic-memory fusion.



We evaluate MarvisNav on standard indoor ZSON benchmarks, where it achieves strong performance without policy training. It also outperforms representative VLM-based methods with far fewer VLM queries, demonstrating practical efficiency for long-horizon navigation. We further validate its practical deployability on a Go2 robot across diverse scenes. Beyond MarvisNav, our study further demonstrates that exploration-memory representation can substantially affect both VLM preferences and navigation performance. These findings highlight the need to consider not only what an agent remembers, but also how that memory is made visible and associated with its choices.

\section{Related Work}

\paragraph{Semantic Guidance and Spatial Choices in ZSON.}

ZSON asks an agent to find an object in an unseen environment without category-specific training. A central problem in ZSON is determining where the target is likely to be found. Early approaches transfer pretrained vision-language knowledge to navigation through multimodal goal embeddings and open-vocabulary localization~\citep{zson, CoWs}. Later methods score the similarity between observations and the target with a pretrained image-text model~\citep{clip} and accumulate it in maps to guide exploration toward promising areas~\citep{vlfm,apexnav,openfmnav}. Large-language models (LLMs) extend this guidance beyond visual similarity by reasoning about object-room relations, scene context, and spatial structure \citep{esc,sgnav,cognav}. VoroNav follows this line and uses a Voronoi Graph to organize navigable paths and sparse observation locations for language-model reasoning \citep{voronav}. More recently, generative VLMs have moved this reasoning onto the egocentric image, judging which directions look promising without an intermediate textual description~\citep{wmnav,DRIVE-Nav}. This shift requires spatial alternatives to be presented in a form that the VLM can visually inspect and select~\citep{som, PIVOT}, while remaining tied to navigation goals that the agent can execute. Recent works provide such an interface by rendering frontiers directly into the observation as selectable navigation goals \citep{openfrontier}. However, these visual choices indicate where the agent can go, but not whether unexplored space remains beyond each option, preventing direct joint reasoning over semantic promise and exploration state for VLM.

\paragraph{Exploration Memory in VLM-Based Object Navigation.}

A VLM query does not retain exploration state across successive decisions, so VLM-based ZSON methods maintain explicit memory to avoid repeatedly considering already searched areas. Some methods keep memory outside the VLM's semantic reasoning: exploration history may restrict the candidates before querying VLMs~\citep{DRIVE-Nav,openfrontier}, or adjust its preference afterward through fixed filtering or score modification~\citep{wmnav,instructnav}. In both cases, semantic evidence and exploration state are evaluated separately and combined by an external rule, placing part of the decision outside the VLM and constraining its judgment. A more direct alternative is to expose the memory to the VLM itself. Text-based methods serialize past observations or actions into the prompt~\citep{History-Augmented,wu2026rememnav,dynavlm} for VLM reasoning. However, the VLM must implicitly associate this linguistic history with the current visual choices, which can make route-memory correspondence ambiguous. Rendering exploration history into a top-view map reduces that alignment problem by placing candidates and explored regions in a shared coordinate frame~\citep{topvnav}. However, reasoning directly over such a map moves the VLM away from the egocentric visual domain in which its perception and semantic reasoning are most naturally trained~\citep{wmnav}. A further step is to bring memory into the VLM's egocentric view itself. Mem2Ego~\citep{mem2ego} projects candidate choices and visited locations into the view as separate markers. This preserves egocentric reasoning, but the correspondence between a candidate and its visitation memory must still be inferred from their image-space proximity. Moreover, a visited marker indicates where the agent has been, but not whether unexplored space remains beyond that location. Our work instead grounds topological exploration memory explicitly onto visual choices, indicating whether unexplored connected space remains beyond each choice. This allows the VLM to jointly reason over semantic evidence and exploration state, without recovering correspondences across representations or relying on external rules.

\section{Motivation: How Memory Representation Shapes Decisions?}
\label{sec:motivation}

\begin{figure}[t]
    \centering
    \includegraphics[width=0.90\linewidth]{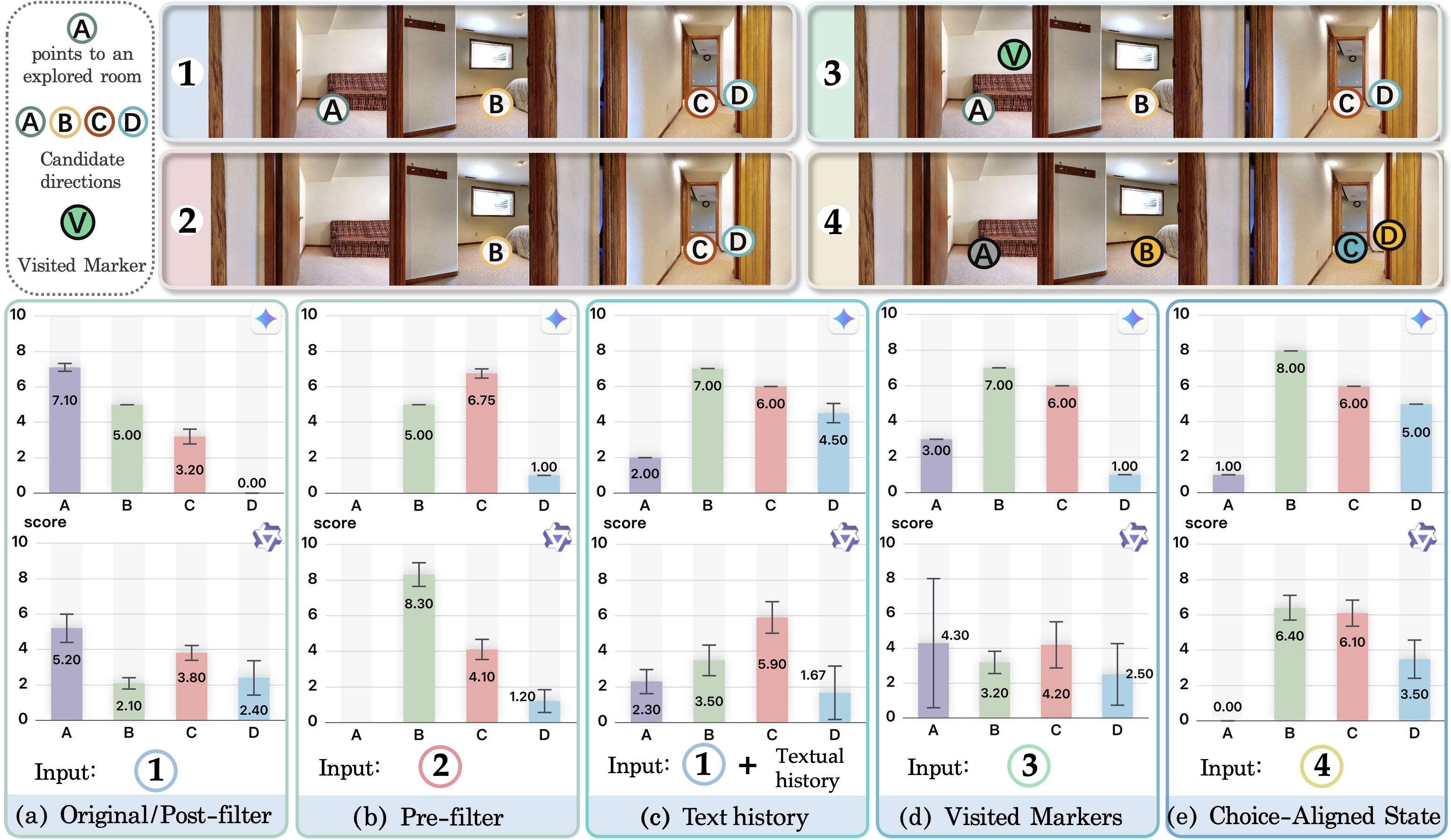}
    \caption{\textbf{Motivating comparison of memory representations.} Gemini-3.5-Flash and Qwen3.5-Plus score each candidate from 0-10 (higher is more worth exploring), with mean and standard deviation over 10 runs. Full prompts and representation-specific changes are detailed in Appendix~\ref{motivation_details}.}
    \label{motivation}
\vspace{-0.2cm}
\end{figure}

Before introducing our method, we conduct a motivating study of
how the representation of exploration memory shapes VLM decisions in ZSON.
Generative VLM-based methods choose where to explore by evaluating visual
navigation choices, making memory representation part of the decision context. While prior methods represent exploration history in diverse ways, comparatively less attention has been paid to how the representation itself affects VLM decision-making. To study this effect in a controlled setting, we keep the scene and target fixed and vary only how exploration memory is represented, as illustrated in Fig.~\ref{motivation}. The five representations span progressively tighter integration between exploration memory and the VLM's visual choices. \textbf{(a) Original / Post-filter} provides the reference VLM scores without exploration memory. It also represents the VLM's original judgment in post-filtering methods, where exploration memory
is applied only after the VLM's evaluation~\citep{wmnav}.
\textbf{(b) Pre-filtering} instead removes an explored candidate before VLM
evaluation~\citep{openfrontier}. \textbf{(c) Textual History} provides prior navigation experience as linguistic context~\citep{History-Augmented}. \textbf{(d) Visited Markers} makes visitation memory visible in the egocentric view but represents it separately from the route candidates~\citep{mem2ego}. \textbf{(e) Choice-Aligned State} is a simplified version of our method, which directly attaches the relevant exploration state to each candidate, making the memory-choice correspondence explicit.


The results show that changing the memory representation can substantially
alter both route scores and rankings. The two external filtering schemes in (a,b) reveal two distinct effects. Under post-filtering, the explored candidate is removed only after VLM evaluation, leaving the final choice to the residual ranking without reconsidering the remaining routes. More counterintuitively, removing candidate \textit{A} before evaluation changes the VLM's relative preferences among the remaining choices: Gemini switches its top choice from \textit{B} to \textit{C}, while Qwen switches from \textit{C} to \textit{B}. This suggests that VLM preferences are sensitive to the composition of the candidate set.  Providing memory directly to the VLM introduces a different challenge: its correspondence with the current visual choices may remain implicit. With \textbf{Textual History} in (c), Qwen assigns the explored candidate \textit{A} a higher mean score than the unvisited \textit{D} ($2.30$ vs.~$1.67$), showing that textual history may not reliably translate into the intended route preference. \textbf{Visited Markers} in (d) projects visited locations into the view makes exploration memory more direct to the VLM, but the marker and candidate remain separate visual elements. Gemini likewise scores explored candidate \textit{A} above unvisited \textit{D}, while Qwen shows less stable and higher variance across repeated queries, resulting unreliable judgment. In contrast, \textbf{Choice-Aligned State} in (e) makes the memory-choice correspondence explicit by attaching exploration state directly to each candidate. Both models clearly separate the explored candidate \textit{A} from the unexplored alternatives. Notably, Qwen produces more differentiated scores and a clearer ranking with choice-aligned states than with separate visitation markers, with explored A scored 0. 

These observations show that the representation of exploration memory directly influences how a VLM evaluates navigation choices. Making memory available, or even visually visible, is not sufficient when its correspondence to the current route remains implicit. This motivates our MarviNav to make topological exploration memory both visible and choice-aligned, by grounding the relevant exploration state explicitly onto the egocentric route choice to which it applies. Rather than indicating only where the agent has been, each choice conveys whether useful unexplored space may remain beyond it, allowing the VLM to jointly reason about semantic promise and exploration state.

\section{Method}

\paragraph{Task formulation.}
We consider ZSON, in which the agent receives no category-specific navigation training for the target categories evaluated at test time. Each episode starts at pose $\mathbf{p}_0$ in an unknown scene with a language-specified target category $c$.
At each step $t$, the agent receives its pose $\mathbf{p}_t$ and an egocentric RGB-D observation $\mathcal{O}_t=\{\mathbf{I}_t,\mathbf{D}_t\}$, where $\mathbf{I}_t$ and $\mathbf{D}_t$ are the RGB and depth images, and selects an action $\mathbf{a}_t$.
An episode succeeds if the agent issues \textsc{stop} at $t\leq T$ within distance $d_s$ of an instance of $c$, where $T$ is the episode step budget.

\paragraph{Overview.}
MarvisNav turns continuous exploration into sparse high-level decisions, as shown in Fig.~\ref{fig2}. As the agent moves, it integrates depth observations into a global metric grid map and incrementally constructs a state-aware RVG whose junctions, endpoints, and connecting paths represent traversable space. Junctions serve as sparse decision sites.
At each site, visible paths and their graph continuations define candidate routes, whose exploration states are determined by frontier association and local graph connectivity. MarvisNav projects the relevant paths and candidate nodes into topology-adaptive RGB views, annotating each candidate with a decision-local letter and a color that encodes its exploration state. These annotations form memory-bearing visual route choices that present scene semantics and exploration state together.
The VLM selects a letter that maps to the corresponding RVG node and specifies a metric navigation goal for the classical navigation stack. After the goal is executed, new observations update the metric grid map, RVG, and exploration states, and the updated graph supplies the next decision site.
This loop continues until the target is found or the step budget is exhausted.

\begin{figure}[t]
    \centering
    \includegraphics[width=1\linewidth]{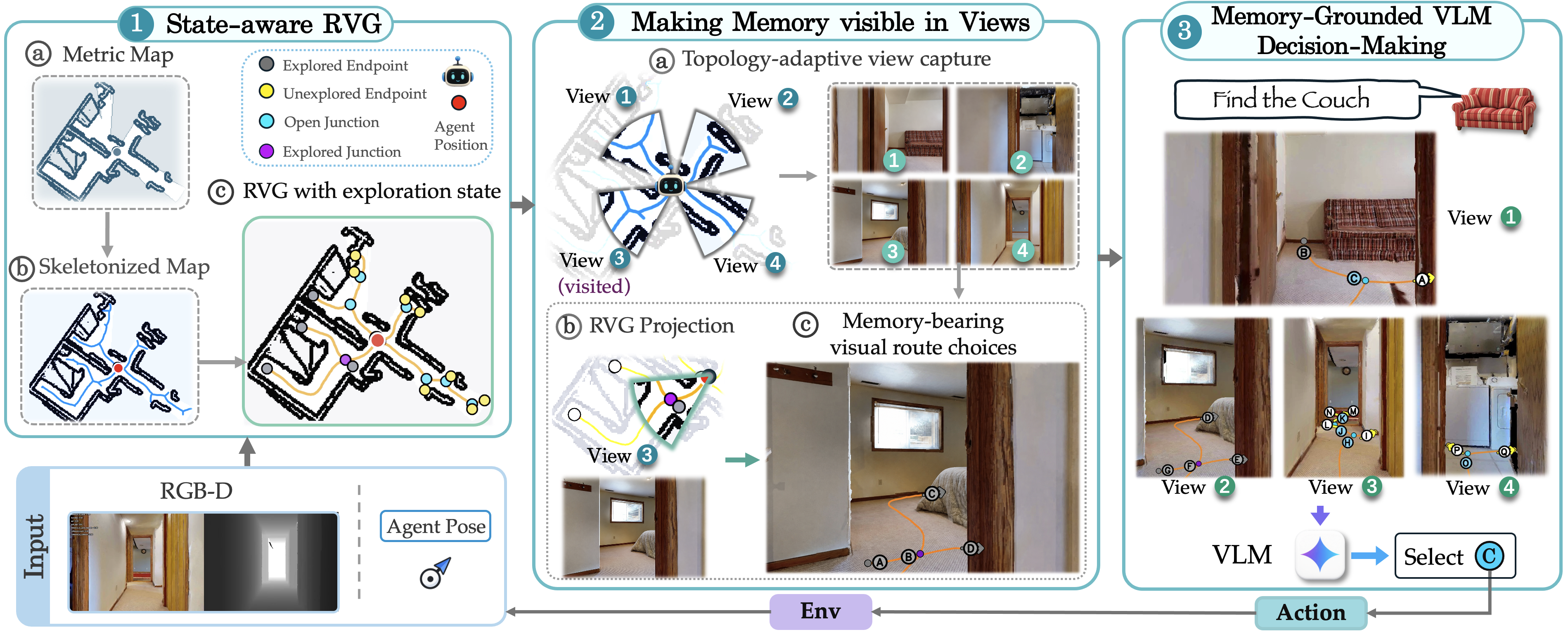}

\caption{
\textbf{Overview of MarvisNav.}
(1) Given RGB-D observations and the agent pose, MarvisNav constructs a state-aware RVG by skeletonizing navigable space and assigning each node a local exploration state.
(2) At each decision site, the local RVG topology determines the number and orientation of egocentric views. Visible paths and candidate nodes are projected into these views, where each candidate is annotated with a letter label and a color-coded exploration state.
(3) Given the target and annotated views, a VLM jointly evaluates target relevance and exploration state. Its selected letter directly determines the next navigation goal.
}
    \label{fig2}
\vspace{-0.2cm}
\end{figure}

\subsection{State-aware Reduced Voronoi Graph}
MarvisNav maintains an online state-aware RVG that compresses observed free space into a sparse topology for route-level exploration.
Following VoroNav~\citep{voronav}, we first construct an RVG in which junctions provide decision sites and graph paths define route choices.
We then augment its nodes with exploration states derived from frontier association and local graph connectivity, allowing the graph to capture local exploration progress beyond binary visitation.

\paragraph{RVG Construction.}
At step $t$, we integrate the depth image $\mathbf{D}_t$ using the associated pose $\mathbf{p}_t$ to maintain a global 2D grid map $\mathbf{M}_t$.
Each grid cell is labeled as free, occupied, or unknown, and observations accumulated earlier in the episode are retained.
Following ApexNav~\citep{apexnav}, we remove mapping artifacts and keep the free-space region that is reachable from the agent.
We skeletonize this region and construct the RVG $\mathcal{G}_t=(\mathcal{V}_t,\mathcal{E}_t)$, where the vertex set $\mathcal{V}_t=\mathcal{V}_t^{\mathrm{junc}}\cup\mathcal{V}_t^{\mathrm{end}}$ contains junction and endpoint vertices, and the edge set $\mathcal{E}_t$ contains the traversable skeleton paths that connect them. We refer to the elements of $\mathcal{V}_t$ as RVG nodes throughout the paper.
Skeleton points with degree one define endpoint nodes, whereas points with degree three or greater define junction nodes.
We trace the paths between these vertices, merge nearby junctions, and prune short terminal branches. The graph is updated online as newly observed free space changes the map.
We also derive a frontier mask from $\mathbf{M}_t$ by identifying the boundary between observed free space and unknown space. Associating this mask with the graph provides the frontier information used to determine each node's exploration state.


\paragraph{Topology-conditioned exploration state.}
Rather than representing exploration memory with a binary visited or unvisited label, we assign each RVG node a topology-conditioned state that summarizes its local exploration status.
Visitation alone is insufficient because a visited junction may still expose an unexplored branch, while an unvisited endpoint may no longer offer a useful exploration direction.
We define the exploration-state set as $\mathcal{S}=\{S_{\mathrm{ue}},S_{\mathrm{ee}},S_{\mathrm{ij}},S_{\mathrm{lj}}\}$ and assign every node $v\in\mathcal{V}_t$ exactly one state through the mapping $\sigma_t:\mathcal{V}_t\rightarrow\mathcal{S}$. Augmenting the RVG with this mapping yields the state-aware graph $\widetilde{\mathcal{G}}_t=(\mathcal{V}_t,\mathcal{E}_t,\sigma_t)$.
The four state values match the visual legend in Fig.~\ref{fig2}. $S_{\mathrm{ue}}$ denotes an \emph{unexplored endpoint} associated with an active frontier, whereas $S_{\mathrm{ee}}$ denotes an \emph{explored endpoint} with no remaining frontier association. $S_{\mathrm{ij}}$ denotes an \emph{open junction} with at least one directly connected node associated with an active frontier, whereas $S_{\mathrm{lj}}$ denotes an \emph{explored junction} with no such neighbor.
These states summarize exploration progress in the local graph neighborhood and provide the topological memory used for subsequent route decisions.

\subsection{Grounding Topological Memory in Egocentric Views}

\label{sec:egocentric_grounding}

Given the current state-aware RVG $\widetilde{\mathcal{G}}_t$, MarvisNav grounds topological exploration memory in the agent's egocentric observations. This involves topology-adaptive view capture, visibility-aware graph projection, and the construction of memory-bearing visual route choices.

\paragraph{Topology-adaptive view capture.}
Let $d$ index the high-level decisions made by the VLM. At decision $d$, which occurs at the current navigation step $t$, MarvisNav captures RGB-D views oriented toward the relevant RVG paths around the current decision site.
Paths with substantially overlapping viewing directions share one view. The number of captured views, denoted by $K_d$, therefore adapts to the local graph topology.
We denote the RGB and depth images of view $k$ by $\mathbf{I}_d^k$ and $\mathbf{D}_d^k$, respectively, where $k=1,\ldots,K_d$.
Unlike WMNav~\citep{wmnav}, which constructs a fixed six-view $360^\circ$ panorama at every decision, MarvisNav captures only the views needed to cover the relevant route alternatives. For example, a single view is sufficient when only one new doorway is available beyond the incoming corridor. This topology-adaptive capture excludes walls and other irrelevant directions from the VLM input.

\paragraph{Visibility-aware RVG projection.}
For each captured view, we transform the current RVG nodes and sampled edge points from the world frame into the camera frame and project them using the camera intrinsics.
A projected point is retained only when it falls inside the image and its projected depth agrees with the observed depth $\mathbf{D}_d^k$. This depth-consistency test removes graph geometry occluded by the observed scene.
We overlay the retained path segments and nodes on $\mathbf{I}_d^k$, which grounds the local RVG structure in the visual scene.
If a candidate node $v_i$ is visible, we mark its projected location. If the node lies outside the image while part of its associated path remains visible, we place an arrow where that path exits the image. This arrow denotes the same candidate rather than introducing another one.
Duplicate arrows are merged within a view, and an arrow is omitted when another view at the same decision already represents the corresponding route continuation.

\paragraph{Memory-bearing visual route choices.}
MarvisNav constructs each annotated RGB image $\widetilde{\mathbf{I}}_d^k$ from $\mathbf{I}_d^k$ by assigning every displayed candidate $v_i$ a decision-local letter label $\ell_{d,i}$ and coloring its marker or arrow according to the current exploration state $\sigma_t(v_i)$.
If the candidate appears in multiple views, it retains the same letter throughout decision $d$.
These annotated images form memory-bearing visual route choices because the scene context, executable route, and corresponding exploration state appear together in the same egocentric view.

\subsection{Memory-Conditioned VLM Decision and Goal Execution}

\paragraph{Memory-conditioned VLM decision.}
At each decision site, the VLM receives the annotated images $\{\widetilde{\mathbf{I}}_d^k\}_{k=1}^{K_d}$ and a text prompt $q(c)$ instantiated with the target category $c$. The prompt explains the decision-local letter labels and state colors, instructs the VLM to balance target relevance against remaining exploration value, and specifies the required output. Prompt examples are provided in Appendix~\ref{app:navigation_prompts}.
The VLM selects the letter associated with the route that is both semantically promising for the target and worth exploring, and it also reports whether the target may already be visible:
\begin{equation}
    (\hat{\ell}_d,z_d)
    =
    \operatorname{VLM}
    \left(
        q(c),\{\widetilde{\mathbf{I}}_d^k\}_{k=1}^{K_d}
    \right),
    \label{eq:vlm_decision}
\end{equation}
where $\hat{\ell}_d$ is the selected letter and the scalar $z_d\in\{0,1\}$ indicates whether the target may be visible.
For every choice, the surrounding RGB content provides evidence about where the target may be, while the marker color conveys the exploration state of the corresponding route. Their spatial alignment allows the VLM to use both sources of evidence in one reasoning step, without a separate rule that combines or reranks semantic and memory scores afterward.

\paragraph{Goal execution and target approach.}

The system resolves $\hat{\ell}_d$ through the decision-local label mapping to its candidate node $v_i$ and metric coordinate $\mathbf{x}(v_i)$. If no target detection is accepted, this coordinate becomes the next goal for the classical navigation stack, which handles path planning, collision avoidance, and low-level control.
A separate detector processes incoming observations throughout exploration. A detection above its standard confidence threshold causes the agent to leave route-level exploration and approach the detected target. When $z_d=1$, MarvisNav lowers the detector threshold for the views captured at that decision so that a weaker detection supported by the VLM can also initiate target approach.
If no detection is accepted, the agent continues toward the selected RVG goal. The resulting observations update $\mathbf{M}_t$, $\widetilde{\mathcal{G}}_t$, and the node states until the agent reaches the target and issues \textsc{stop}, or $t=T$.

\section{Experiment}
Our experiments systematically assessed the proposed MarvisNav around the following questions:
\begin{itemize}
    \item How does MarvisNav perform against existing ZSON methods? (\ref{comparison with baselines})
    \item How do different memory representations affect navigation performance? (\ref{Ablation Study on Memory-Grounded Visual Route Choices})
    \item Does MarvisNav remain robust across VLM backbones, and at what query cost? (\ref{VLM Backbone Ablation and Query Efficiency})
    \item Can MarvisNav perform object searches on real robots without task-specific training? (\ref{real-world})
\end{itemize}

We report Success Rate (SR,$\%$) and Success weighted by Path Length (SPL,$\%$) under the standard ObjectNav evaluation protocol. SR measures the fraction of episodes completed successfully, while SPL additionally penalizes successful trajectories that are longer than the shortest optimal path.  
Due to space constraints, we report the real-robot experiments in Appendix~\ref{real-world}.

\begin{table}[t]
\centering
\small
\setlength{\tabcolsep}{5pt}

\caption{\textbf{Comparison with ObjectNav baselines} on HM3D v0.1, HM3D v0.2, and MP3D. ZS denotes zero-shot ObjectNav, without task-specific fine-tuning.  Results are taken from the original papers when available; otherwise, we use reproductions reported in~\citep{apexnav}: HM3D-v0.2(L3MVN, VLFM, and SG-Nav), and MP3D (L3MVN and OpenFMNav).}
\begin{tabular}{llcrrrrrr}
\toprule
 & & & \multicolumn{2}{c}{HM3D v0.1} & \multicolumn{2}{c}{HM3D v0.2} & \multicolumn{2}{c}{MP3D} \\
\cmidrule(lr){4-5}\cmidrule(lr){6-7}\cmidrule(lr){8-9}
Reasoning & Method & ZS & SR$\uparrow$ & SPL$\uparrow$ & SR$\uparrow$ & SPL$\uparrow$ & SR$\uparrow$ & SPL$\uparrow$ \\
\midrule
\multirow{2}{1.8cm}{Policy}

 & PONI~\citep{poni}           & \xmark & --   & --   & --   & --   & 31.8 & 12.1 \\
 & Uni-NaVid~\citep{Uni-NaVid}     & \xmark & --   & --   & 73.7   & 37.1   & -- & -- \\
    
\midrule
\multirow{5}{1.8cm}{Embedding similarity-based guidance}
 & ZSON~\citep{zson}           & \cmark & 25.5 & 12.6 & --   & --   & 15.3 & 4.8  \\
 & CoW~\citep{CoWs}            & \cmark & --   & --   & --   & --   & 7.4  & 3.7  \\
 & VLFM~\citep{vlfm}           & \cmark & 52.5 & 30.4 & 63.6 & 32.5 & 36.4 & 17.5 \\
 & OpenFMNav~\citep{openfmnav} & \cmark & 54.9 & 24.4 & --   & --   & 37.2 & 15.7 \\
   & ApexNav~\citep{apexnav}     & \cmark & 59.6   & 33.0   & 76.2 & 38.0 & 39.2 & 17.8 \\
\midrule

\multirow{5}{1.8cm}{LLM-Based Semantic Reasoning}
 & ESC~\citep{esc}             & \cmark & 39.2 & 22.3 & --   & --   & 28.7 & 14.2 \\
 & L3MVN~\citep{l3mvn}         & \cmark & 50.4 & 23.1 & 36.3 & 15.7 & 34.9   & 14.5   \\
 & VoroNav~\citep{voronav}     & \cmark & 42.0 & 26.0 & --   & --   & --   & --   \\
 & SG-Nav~\citep{sgnav}        & \cmark & 54.0   & 24.9   & 49.6 & 25.5 & 40.2 & 16.0 \\
 & UniGoal~\citep{unigoal}     & \cmark & 54.5 & 25.1 & --   & --   & 41.0 & 16.4 \\

\midrule
\multirow{7}{1.8cm}{Generative VLM-Based Visual Reasoning}
 & TopV-Nav~\citep{topvnav}          & \cmark & 52.0 & 28.6 & --   & --   & 35.2 & 16.4 \\
 &His-aug~\citep{History-Augmented}  & \cmark & 46.0 & 24.8 & --   & --   & -- & -- \\
 & InstructNav~\citep{instructnav}   & \cmark & --   & --   & 58.0 & 20.9 & --   & --   \\
  & WMNav~\citep{wmnav}               & \cmark & 58.1 & 31.2 & 72.2 & 33.3 & 45.4 & 17.2 \\
 & DRIVE-Nav~\citep{DRIVE-Nav}        & \cmark & 55.5 & 35.9 & 72.4 & 41.3 & 41.8 & 22.6 \\
 & RememNav~\citep{wu2026rememnav}   & \cmark & 60.0 & \textbf{36.8} & 67.8 & 36.6 & \textbf{49.8} & \textbf{24.3}   \\
  & OpenFrontier~\citep{openfrontier}   & \cmark &  -- &  --& 77.3 & 35.6 & 40.7 & 17.8   \\
 \rowcolor{oursbg} & \textbf{MarvisNav (ours)}         & \cmark & \textbf{60.5} & 36.2 & \textbf{81.2}& \textbf{42.5} & 42.1 & 20.4 \\
\bottomrule

\end{tabular}

\label{tab:main_results}

\end{table}

\subsection{Comparison with Baselines}
\label{comparison with baselines}
Table~\ref{tab:main_results} compares MarvisNav with representative policy-based, embedding-based, LLM-based, and generative-VLM-based methods. We list trained policies for reference. Among the compared methods, recent generative-VLM approaches form the strongest baselines. Since prior methods use different VLMs reported in their own papers, Sec.~\ref{VLM Backbone Ablation and Query Efficiency} also compares representative methods under a shared backbone. MarvisNav performs particularly well on HM3D v0.2, achieving 81.2\% SR and 42.5\% SPL. These exceed the strongest reported baseline results for each metric by 3.9 percentage points in SR over OpenFrontier and 1.2 points in SPL over DRIVE-Nav. On HM3D v0.1, MarvisNav obtains the highest SR among the compared methods, while its SPL remains close to the best result. It also outperforms Uni-NaVid, which is trained on millions of navigation samples, by 7.5 percentage points in SR. Performance on MP3D is competitive but not uniformly leading. MarvisNav achieves higher SPL but lower SR than WMNav, while ReMemNav outperforms it on both metrics.
Overall, these results show that competitive navigation can be achieved with direct, memory-conditioned VLM selection, without an additional semantic-memory fusion stage.

\subsection{Ablation Study on Memory Representation}
\label{Ablation Study on Memory-Grounded Visual Route Choices}

\begin{wraptable}{r}{0.47\textwidth}
    \vspace{-13pt}                 
    \centering
    \small
    \caption{Ablation of memory representations.}
    \label{tab:memory_ablation}
    \vspace{-6pt}                  
    \setlength{\tabcolsep}{3pt}
    \begin{tabular}{@{}lccc@{}}
        \toprule
        Representation &
        \shortstack{To VLM} &
        SR (\%) $\uparrow$ &
        SPL (\%) $\uparrow$ \\
        \midrule
        No Memory       & --     & 71.9 & 35.4 \\
        Post-filtering  & \xmark & 74.3 & 37.1 \\
        Pre-filtering   & \xmark & 75.0 & 37.2 \\
        Textual History & \cmark & 73.6 & 36.6 \\
        Visited Markers & \cmark & 76.5 & 40.6 \\
        Choice-aligned  & \cmark &
        \textbf{81.2} & \textbf{42.5} \\
        \bottomrule
    \end{tabular}
\end{wraptable}
To evaluate how memory representation affects navigation, we compare six MarvisNav variants on HM3D v0.2 with Gemini-3.5-Flash and the remaining navigation pipeline fixed. All variants share RVG-based candidate generation and egocentric projection of paths and nodes, varying only memory cues and filtering. \textit{No Memory} keeps candidate colors unchanged regardless of exploration status. \textit{Post-filtering} uses this display and applies exploration-based filtering after VLM evaluation, whereas \textit{Pre-filtering} omits explored candidates from the visual overlay before evaluation. \textit{Textual History} supplements the \textit{No Memory} views with summaries of the ten most recent node visits. \textit{Visited Markers} adds green markers beside visited visual candidates, leaving the original candidate markers unchanged. Our \textit{Choice-aligned} representation instead encodes topology-conditioned exploration states directly in candidate markers.

As shown in Table~\ref{tab:memory_ablation}, \textit{Choice-aligned} achieves the highest SR and SPL at $81.2\%$ and $42.5\%$, exceeding \textit{Visited Markers}, the strongest alternative, by $4.7$ and $1.9$ percentage points, respectively. All memory variants improve over \textit{No Memory}, but their benefits differ. In particular, \textit{Textual History} underperforms both external filtering variants, indicating that supplying history directly to the VLM does not necessarily yield greater navigation gains. \textit{Visited Markers} performs better, but presenting topology-conditioned states directly on the candidates provides further improvements. These results support the combined use of topological exploration state and explicit memory-choice alignment, highlighting that effective navigation depends not merely on having memory, but also on how it is represented for current choices.


\subsection{VLM Backbone Ablation and Query Efficiency}
\label{VLM Backbone Ablation and Query Efficiency}

\begin{table}[h]
    \centering
    \begin{minipage}[t]{0.46\linewidth}
        \vspace{0pt}
        \centering
        \caption{Ablation of VLMs on HM3D v0.2.}
        \label{vlm_ablation}
        \begin{tabular}{lcc}
            \toprule
            VLM & SR (\%) $\uparrow$ & SPL (\%) $\uparrow$ \\
            \midrule
            Qwen3-VL-8B      & 76.0          & 38.3          \\
            Qwen3.5-9B       & 76.5          & 37.8          \\
            Qwen3.7-plus     & 77.3          & 39.3
            \\
            Gemma4-26B       & 78.9          & 38.5          \\
            Gemini-2.5-Flash & 78.5          & 40.0          \\
            Gemini-3.5-Flash & \textbf{81.2} & \textbf{42.5} \\
            \bottomrule
        \end{tabular}
    \end{minipage}
    \hfill
    \begin{minipage}[t]{0.48\linewidth}
        \vspace{0pt}
        \centering
        \captionof{figure}{VLM query efficiency on HM3D v0.2.}
        \includegraphics[width=\linewidth]{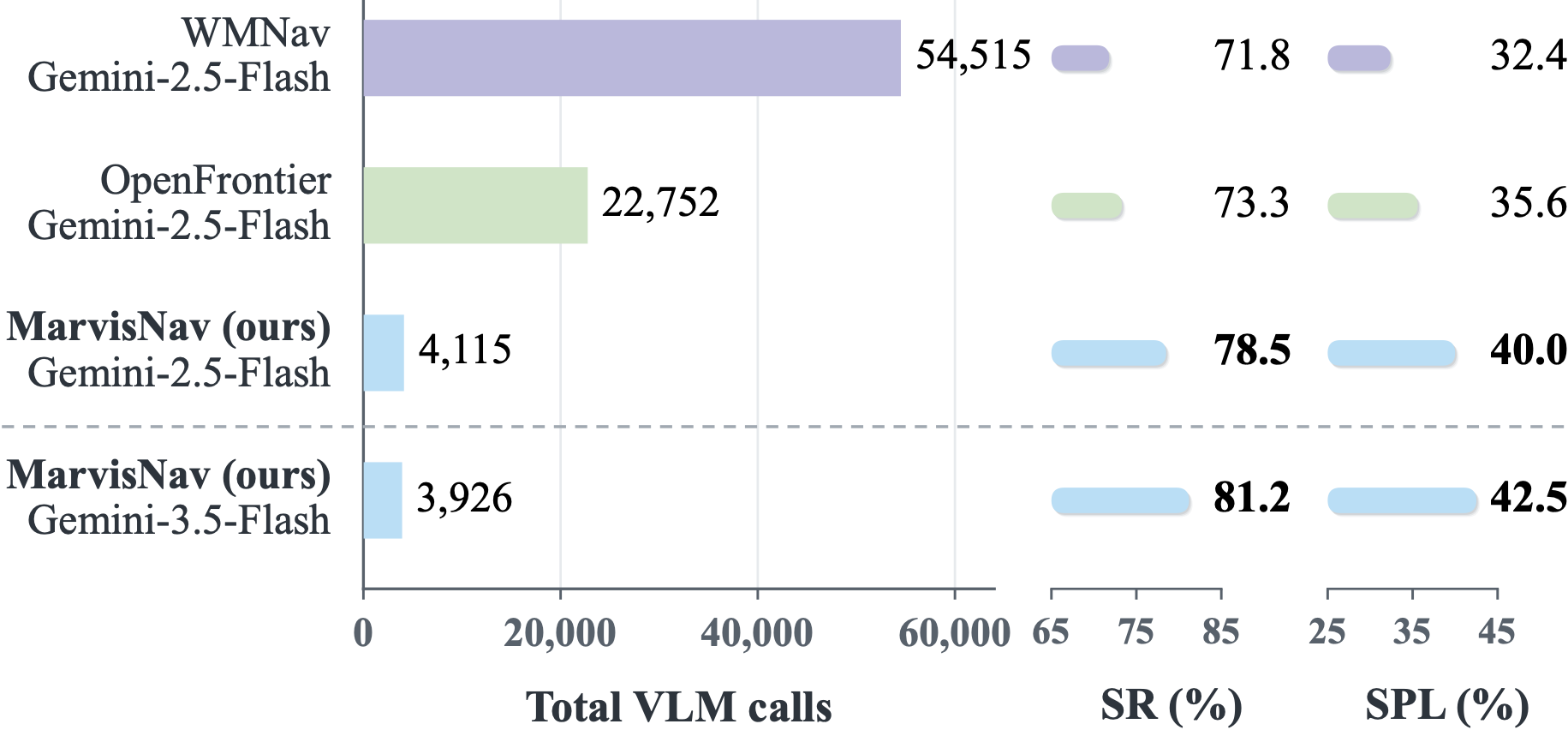}

        \label{query_efficiency}
    \end{minipage}
\vspace{-6pt}
\end{table}

To assess sensitivity to VLM choice, we evaluate MarvisNav with five VLMs on HM3D v0.2 while keeping the rest of the pipeline fixed. As shown in Table~\ref{vlm_ablation}, MarvisNav maintains strong performance across VLMs: even with lighter Qwen3-VL-8B, it achieves $76.0\%$ SR and $38.3\%$ SPL. To separate framework gains from VLM strength, we further compare methods using the same Gemini-2.5-Flash backbone. MarvisNav still outperforms WMNav~\citep{wmnav} and OpenFrontier~\citep{openfrontier} (Fig.~\ref{query_efficiency}), indicating that the gains are not attributable to VLM strength alone, but also to how MarvisNav represents exploration memory and integrates it into the navigation pipeline.
Using Gemini-3.5-Flash further improves performance to $81.2\%$ SR and $42.5\%$ SPL. Overall, MarvisNav retains strong performance with weaker VLMs while continuing to benefit from more capable backbones.



To assess the efficiency of our approach, we compare total VLM queries rather than runtime, as response latency varies with serving conditions. This is important for real-world deployment of VLM-based ZSON, where frequent queries increase navigation waiting time. We reproduce WMNav and OpenFrontier, two representative VLM-based ZSON methods with open-source official implementations, and evaluate all methods using Gemini-2.5-Flash on the same $1{,}000$ HM3D v0.2 episodes. As shown in Fig.~\ref{query_efficiency}, MarvisNav requires $4{,}115$ VLM calls, only $7.5\%$ and $18.1\%$ of those used by WMNav ($54{,}515$) and OpenFrontier ($22{,}752$), respectively, while achieving higher SR and SPL. With Gemini-3.5-Flash, it falls further to $3{,}926$ and success rate reaches $81.2\%$. Notably, higher query frequency does not translate into better navigation performance in this controlled comparison. This query efficiency is consistent with MarvisNav's design: sparse decision sites limit when the VLM is queried, while choice-aligned exploration memory supports more effective decisions.

\section{Conclusion}
\label{sec:conclusion}

We introduced MarvisNav, a framework for VLM-based ZSON that makes topological exploration memory visible on egocentric route choices.
A state-aware reduced Voronoi graph captures local exploration progress beyond binary visitation and organizes navigation into sparse high-level decisions.
By grounding exploration states directly on visual candidates, MarvisNav enables the VLM to jointly evaluate target relevance, exploration state, and make the final route selection without post-hoc semantic-memory fusion.
Experiments on HM3D and MP3D demonstrate strong navigation performance without task-specific policy training, while comparisons under a shared VLM backbone show better performance with substantially fewer queries than representative baselines.
Controlled comparisons and navigation ablations support the benefits of combining topological exploration states with explicit memory-choice alignment.
Real-robot experiments further demonstrate closed-loop object search in the physical world, helping to bridge the sim-to-real gap.
Taken together, these findings show that exploration memory is most effective for VLM-based navigation when its state is clearly represented and directly aligned with the agent's available choices.

\bibliography{iclr2027_conference}
\bibliographystyle{iclr2027_conference}

\newpage
\appendix





\section{Benchmark Evaluation Setup}

\paragraph{Evaluation Protocol and Implementation.}We follow the official protocol without modification, including $640\times480$ RGB-D observations, a $79^\circ$ horizontal field of view, and a 500-step limit. The agent operates in the discrete action space
$\{\textsc{Stop}, \textsc{MoveForward}, \textsc{TurnLeft}, \textsc{TurnRight}, \textsc{LookUp}, \textsc{LookDown}\}$,
with a forward step size of $0.25\,\mathrm{m}$ and a rotation increment of $30^\circ$. We adopt the low-level planning and target-approach modules following ApexNav~\citep{apexnav}. For target perception, we use RF-DETR~\citep{robinson2026rfdetr} and SAM3~\citep{sam3}. The main experiments use Gemini 3.5 Flash for vision-language reasoning. All simulation experiments are run on a desktop equipped with a single NVIDIA GeForce RTX 4090 GPU.

\paragraph{Benchmarks.}We evaluate MarvisNav on three Habitat~\citep{habitat} ObjectNav benchmarks constructed from real-world 3D scans: HM3D-Semantics v0.1, HM3D-Semantics v0.2, and Matterport3D (MP3D) \citep{ramakrishnan2021hm3d,hm3d0.2,mp3d}. HM3D v0.1, used in the Habitat 2022 ObjectNav Challenge, contains 2,000 validation episodes across 20 scenes and six target categories. HM3D v0.2 expands and corrects the semantic annotations, with 1,000 validation episodes across 36 scenes and the same six categories. MP3D provides 2,195 validation episodes across 11 scenes and 21 categories. In total, we evaluate on 5,195 episodes.

\section{Real-World Deployment}
\label{real-world}

\paragraph{Robot platform and navigation setup.} We deploy MarvisNav on a Unitree Go2 to demonstrate closed-loop object search beyond simulation. An Odin1 sensing module provides RGB-D observations and pose estimates, with the camera mounted $0.45\,\mathrm{m}$ above the ground. We limit the depth observation range to $4\,\mathrm{m}$ for both mapping and local navigation. All MarvisNav components except VLM inference run onboard an NVIDIA Jetson AGX Orin with Ubuntu 20.04 and ROS Noetic. Target detection and instance segmentation use YOLO26m-seg~\citep{yolo26}, while Gemini-3.5-Flash is accessed through a remote API for high-level route selection.  We retain MarvisNav's topological memory and visual decision pipeline, replacing the classical map-based planner with goal-conditioned visual navigation~\citep{SRU, wang2026sand}. Specifically, we use the pretrained SRU \citep{SRU} policy without retraining or fine-tuning to navigate toward the metric goals specified by the VLM-selected route identifiers.

\paragraph{Deployment protocol and demonstrations.} We deploy MarvisNav on a Unitree Go2 across diverse real-world environments, including a large office, a home, and outdoor settings, with runs covering daytime and nighttime conditions. Each run starts with an empty map and only a target category, without prior knowledge of the scene layout or target location. The robot performs online mapping, route selection, and navigation without human intervention. Full indoor and outdoor deployment videos are shown in the supplementary material.

Figure~\ref{real_robot} presents a representative indoor run in which the robot searches for a \textit{potted plant in the reception} in a large open-plan office. The robot travels $39.2\,\mathrm{m}$ and records a mapped area of $238.6\,\mathrm{m}^2$. The run includes nine high-level decision and scan sites, O1--O9, with one VLM query per site. At O1, MarvisNav captures six egocentric views for panoramic coverage of the surroundings before route selection. The VLM combines the instruction's reception-area cue with visible office and corridor features to guide context-aware route selection. During exploration, reaching the selected node in the State-Aware RVG triggers the next scan and decision. The plant first becomes visible at O9, after which the robot approaches the target and stops automatically upon arrival.

The point-cloud overview at the top of Fig.~\ref{real_robot} shows the actual robot trajectory, while Fig.~\ref{full_rvg} presents the final State-Aware RVG. Both maps are constructed from observations collected during navigation. The State-Aware RVG represents observed free space through junctions, boundary nodes, and connecting paths, with node states capturing local exploration progress. Some projected path segments on egocentric images appear discontinuous because of depth noise, occlusion filtering, and the $4\,\mathrm{m}$ observation range. The robot nevertheless completes the search, demonstrating closed-loop route selection and execution despite these local projection gaps. The trajectory overlaid in red on the State-Aware RVG shows how the selected visual route choices translate into physical navigation.

\begin{figure*}[t]
    \ContinuedFloat
    \centering
    \includegraphics[width=\textwidth]{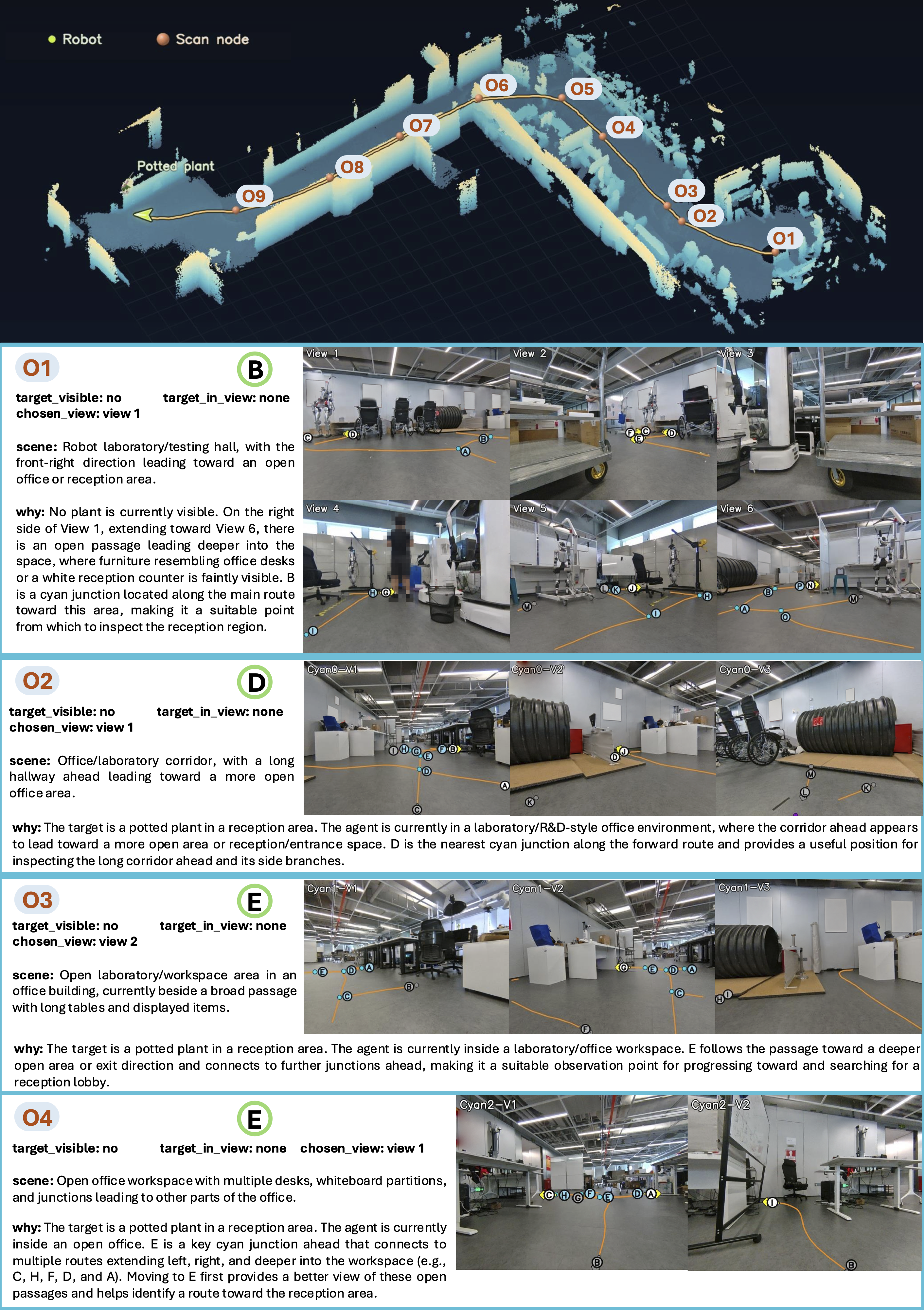}
\end{figure*}
\clearpage
\begin{figure*}[t]
    \centering
    \includegraphics[width=\textwidth]{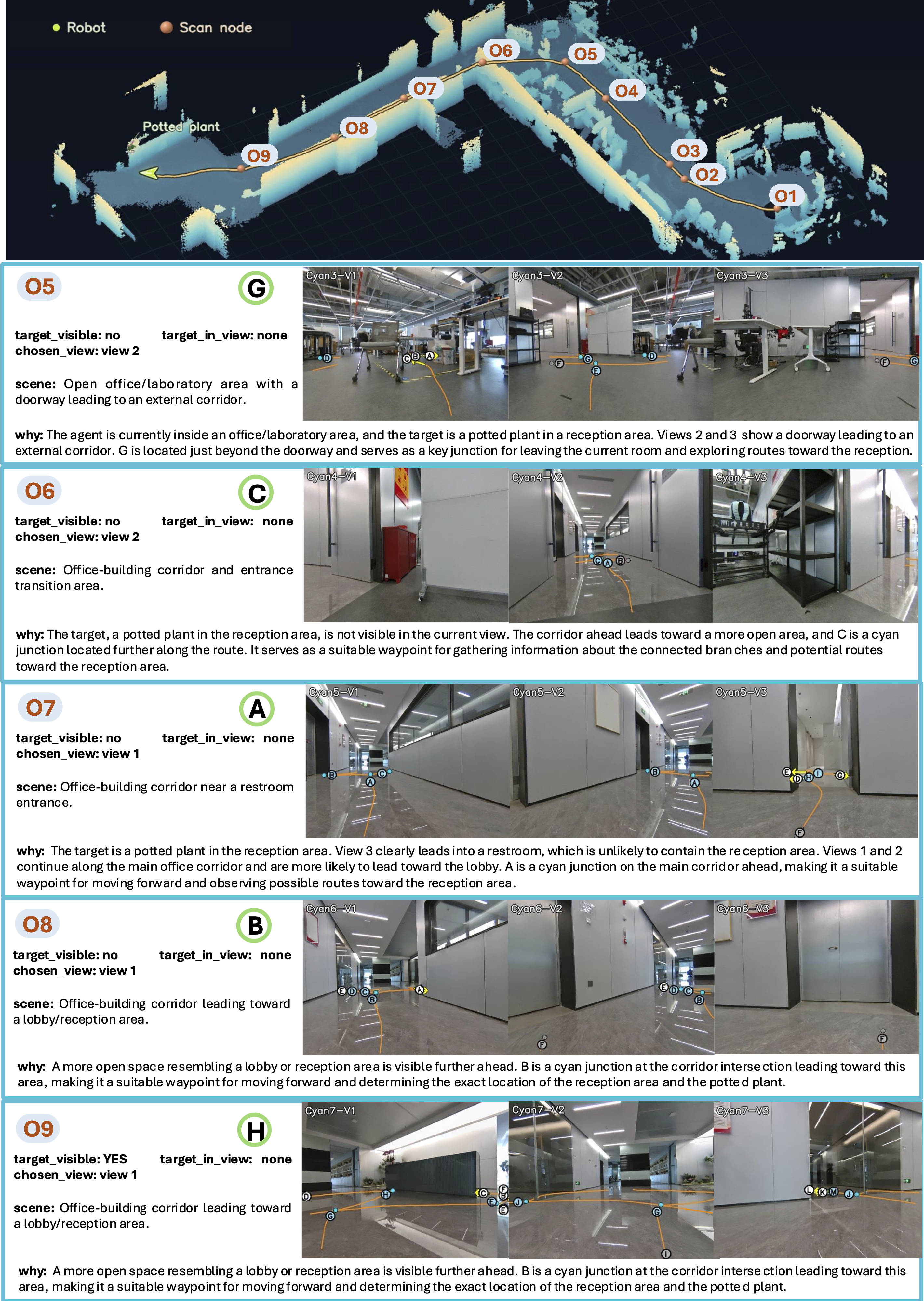}
    \caption{\textbf{Real-world deployment of MarvisNav on a Unitree Go2.} The robot searches for a \textit{potted plant in the reception}. The point-cloud overview shows the executed trajectory and decision sites O1--O9. The accompanying panels present the annotated egocentric views, selected route identifiers, and original VLM responses. Route identifiers are local to each decision.}
    \label{real_robot}
\end{figure*}

\clearpage
\begin{figure}
    \centering
    \includegraphics[width=1\linewidth]{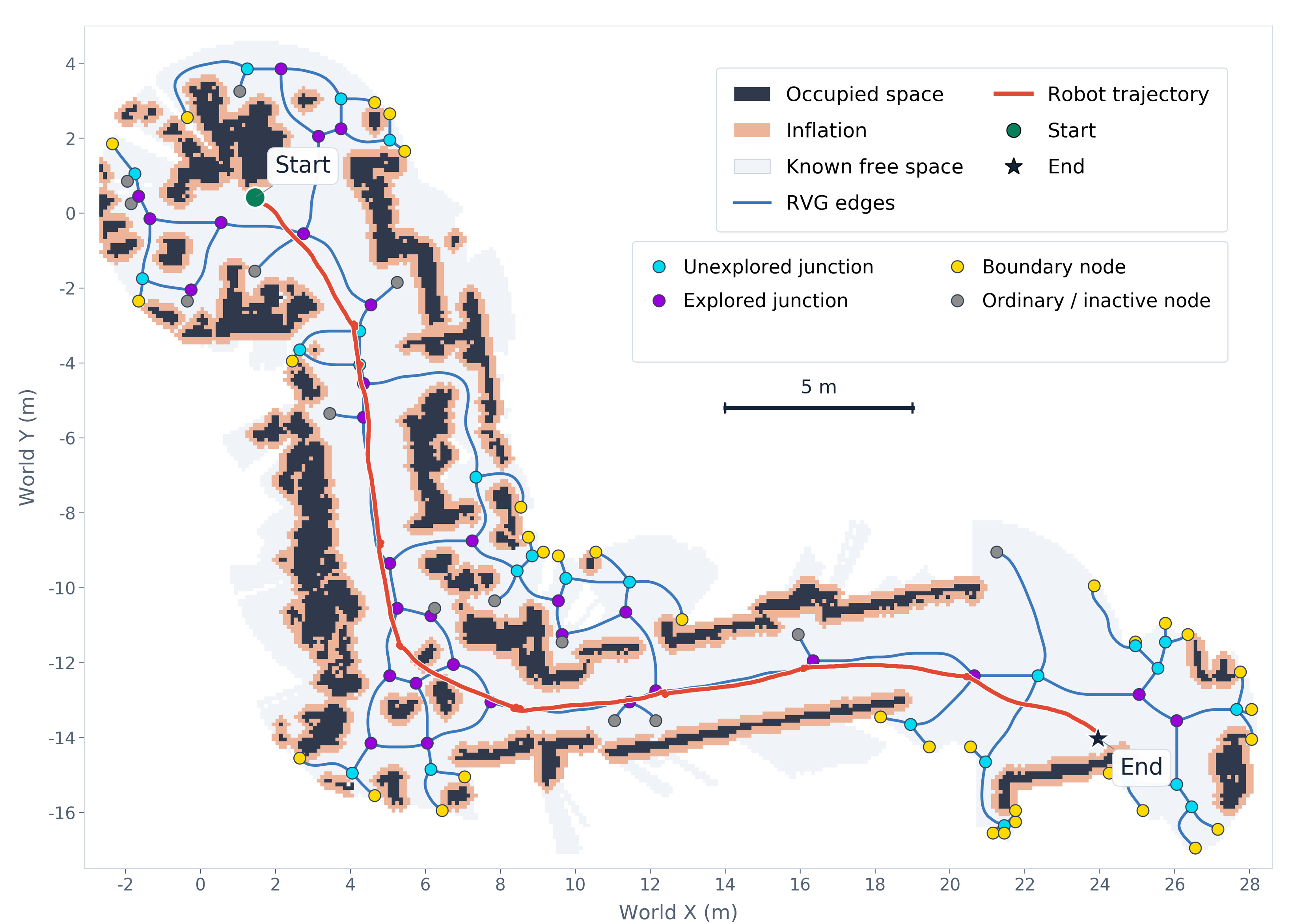}
   \caption{\textbf{State-Aware RVG and executed trajectory in the representative indoor deployment.}
Starting from an empty map, MarvisNav incrementally constructs the RVG and updates node exploration states from online RGB-D observations. The final map shows the resulting topology, exploration states, and robot trajectory from start to end.}
    \label{full_rvg}
\end{figure}

\section{Scene Reconstruction for Visualization}

\label{scene_reconstruction}
\begin{figure*}[h]
    \centering
    \begin{subfigure}[t]{0.32\textwidth}
        \centering
        \includegraphics[width=\linewidth]{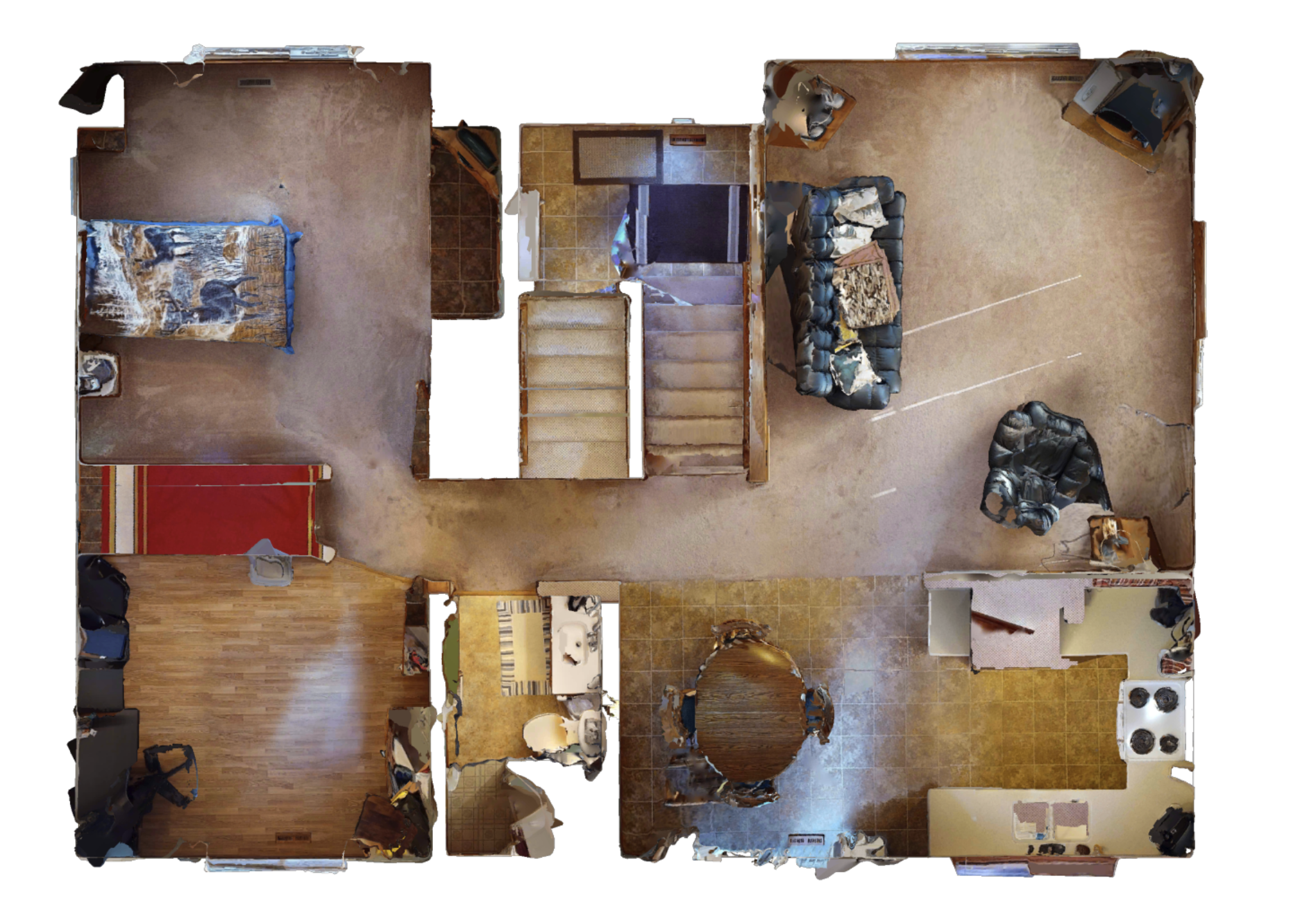}
        \caption{Original HM3D top-down view.}
        \label{fig:hm3d_original}
    \end{subfigure}
    \hfill
    \begin{subfigure}[t]{0.36\textwidth}
        \centering
        \includegraphics[width=\linewidth]{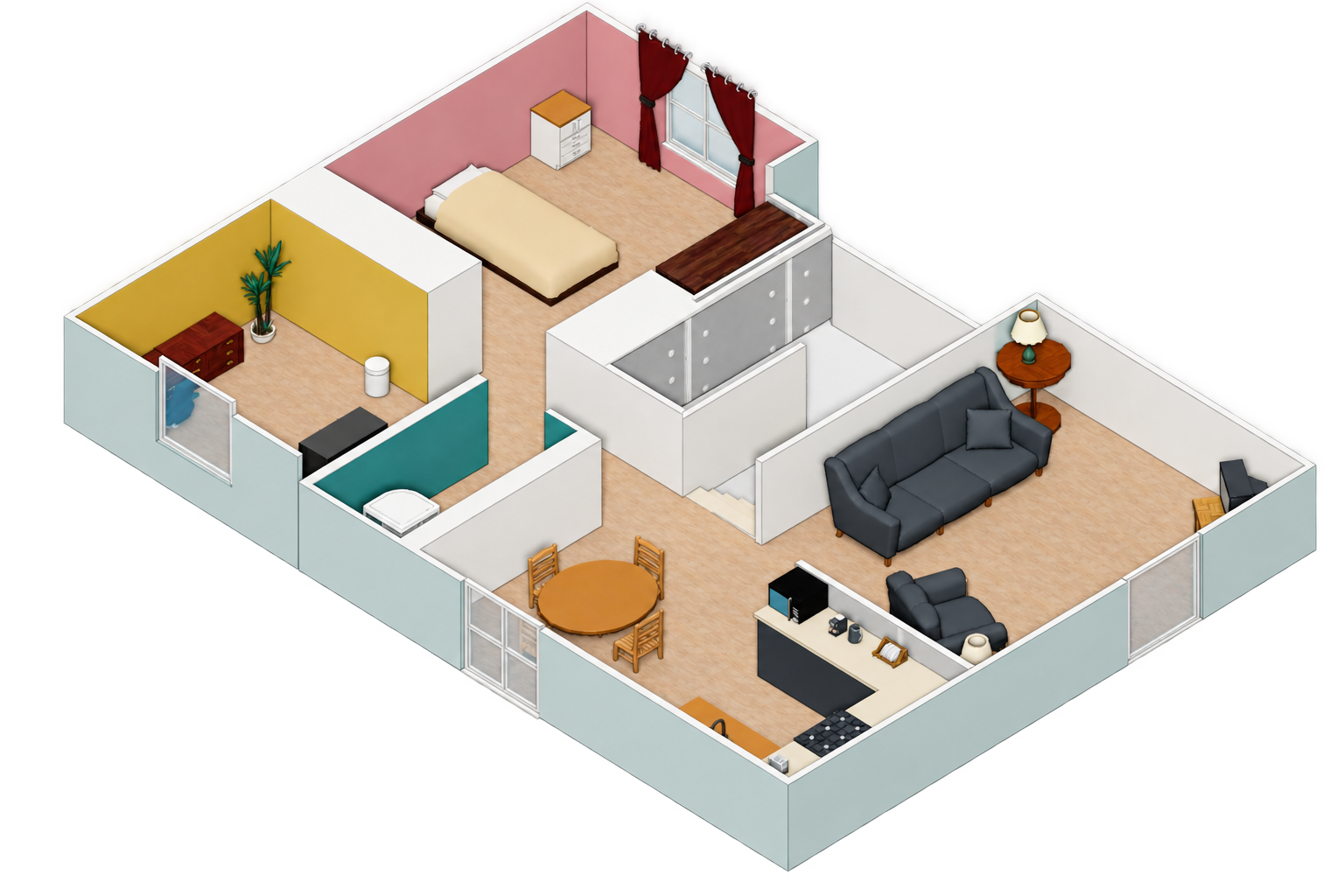}
        \caption{Reconstructed 3D model.}
        \label{fig:hm3d_reconstruction}
    \end{subfigure}
    \hfill
    \begin{subfigure}[t]{0.3\textwidth}
        \centering
        \includegraphics[width=\linewidth]{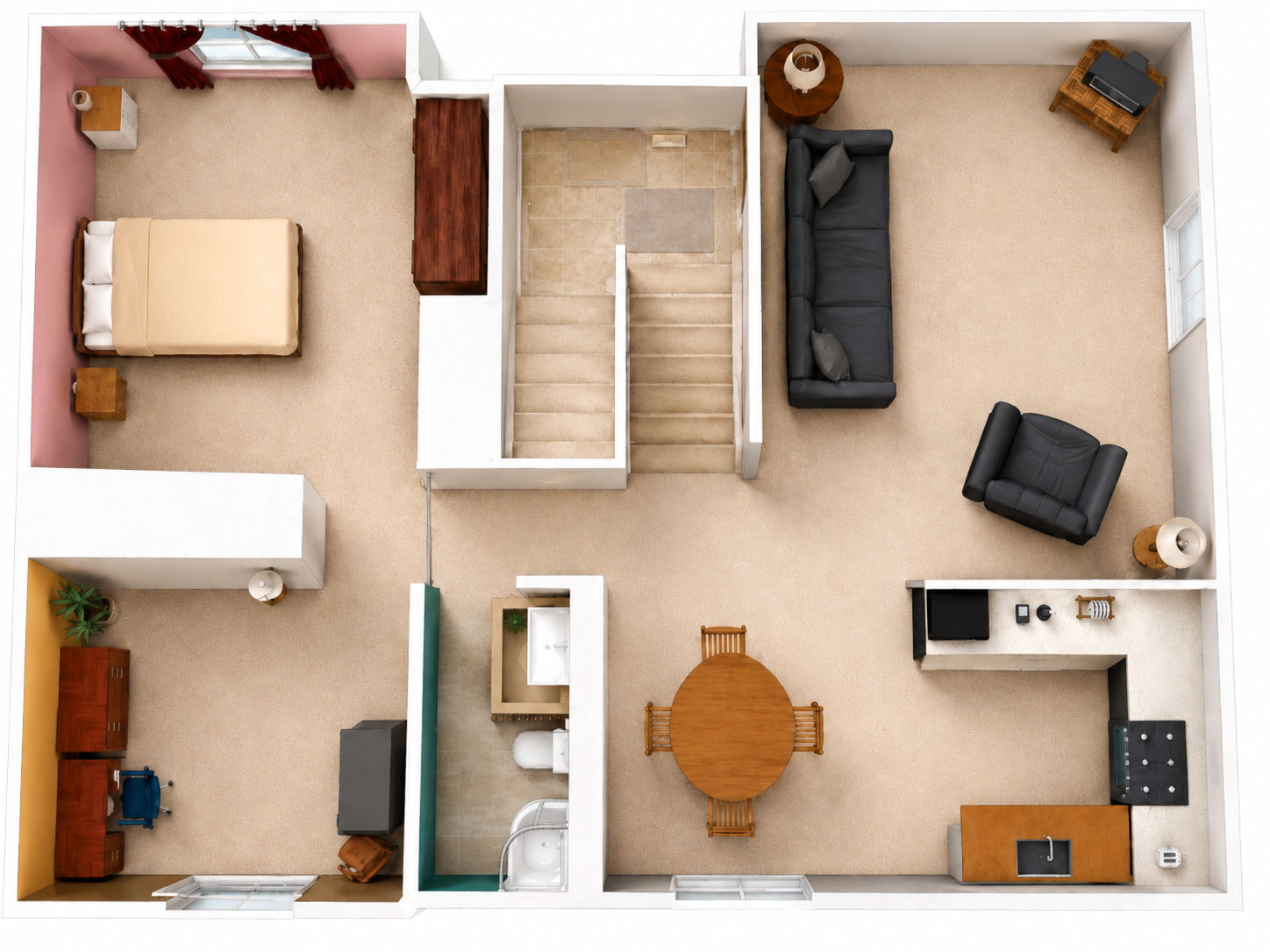}
        \caption{Top-down rendering of the reconstructed model.}
        \label{fig:hm3d_rendered}
    \end{subfigure}

    \caption{
        Scene reconstruction used to create method illustrations.
        The reconstruction follows the layout and relative spatial proportions of HM3D scene \texttt{00877-4ok3usBNeis}.
    }
    \label{fig:hm3d_scene_reconstruction}
\end{figure*}

As shown in Fig.~\ref{fig:hm3d_scene_reconstruction}, raw scan-derived visualizations of HM3D scenes can be visually cluttered, making their underlying spatial layout difficult to parse at a glance. To facilitate interpretation of our method, we independently reconstructed the layout of one HM3D scene using Pascal Editor, following its original layout and relative spatial proportions. This is an actual scene used in our evaluation, rather than a toy environment created solely for illustration. As illustrated in the figure, we present the original top-down view, the reconstructed 3D model, and its rendered top-down view. Importantly, this reconstruction is used only for visualization, and evaluations are performed in the original HM3D environment.  We will release the reconstructed 3D model as a reusable asset for future work, supporting visualization and qualitative analysis in the corresponding HM3D scene.

\section{Details of the Motivating Study}
\label{motivation_details}

\begin{figure}[t]
    \centering
    \includegraphics[width=1\linewidth]{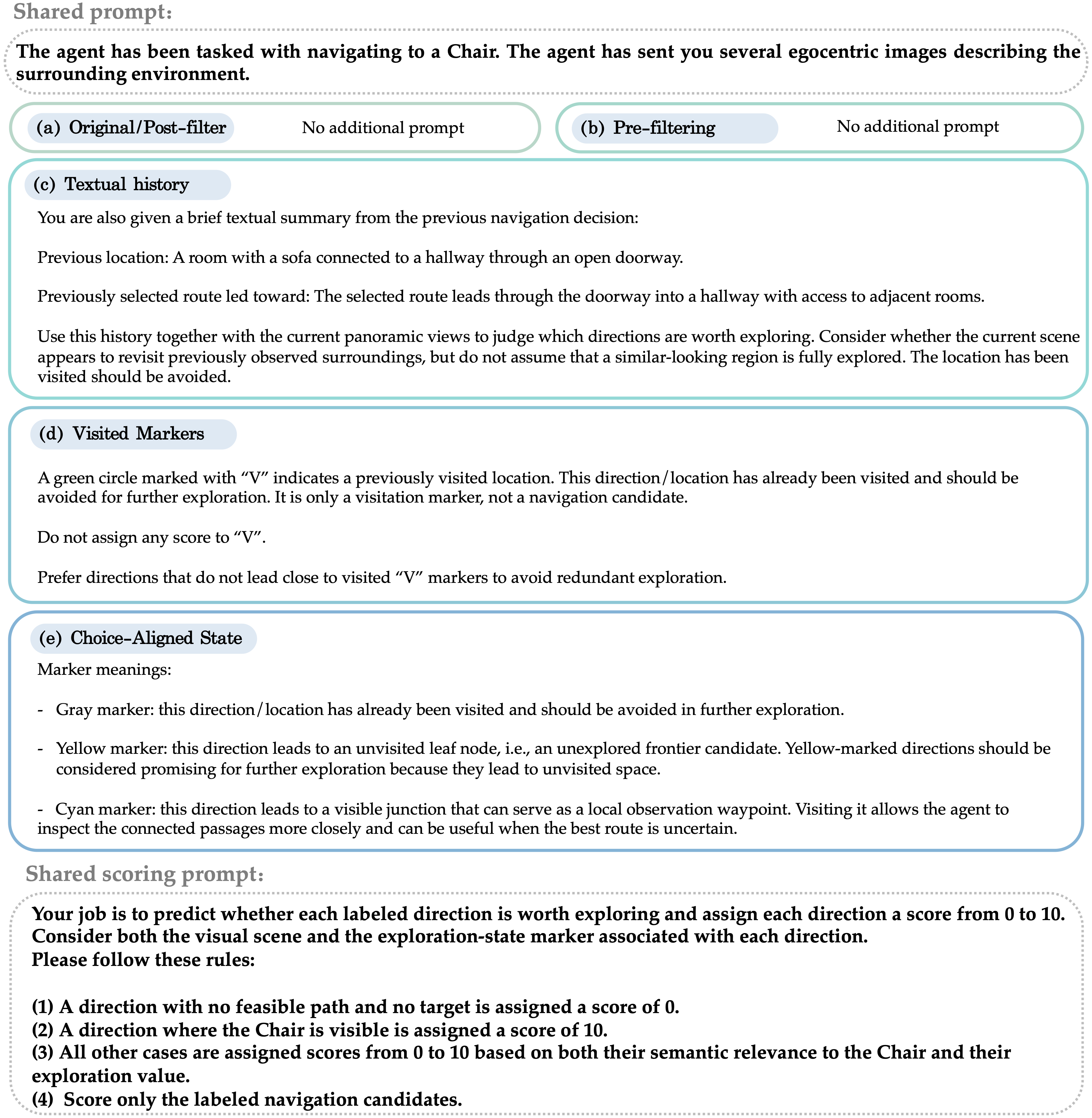}
    \caption{\textbf{Prompt templates for the motivation study.} All variants share the same task and scoring instructions and differ only in how exploration memory is provided.}
    \label{prompt_toy}
\end{figure}

\subsection{Setup and memory representations}

We conducted the motivating study of how memory representation affects the evaluation of the VLM route. Across variants, we fix the navigation scene, target category (\emph{Chair}), underlying egocentric observation, candidate geometry, image resolution, base prompt, and model configuration, making only the changes required by each memory representation. For a simple and visually transparent comparison, we omit the full RVG paths and retain only the candidate markers and memory cues. We compare five representations. \textbf{(a) Original / Post-filter} provides the reference VLM scores without exploration memory; the same raw scores are used before memory is applied by a post-filtering rule. \textbf{(b) Pre-filtering} removes the explored candidate \textbf{A} before VLM evaluation, while \textbf{B}, \textbf{C}, and \textbf{D} retain their original identities and locations. \textbf{(c) Textual History} provides a VLM-generated summary of the previous decision as linguistic memory. For this diagnostic study, we use only the immediately preceding history entry, which describes the location just visited and is directly relevant to the current decision. This provides the minimal setting in which the VLM must associate textual memory with the corresponding visual choice, without introducing additional effects from history length. Our full simulation ablation instead uses the most recent ten entries. \textbf{(d) Visited Markers} places a separate green \texttt{V} marker beside the explored candidate \textbf{A}. \textbf{(e) Choice-Aligned State} attaches the exploration state directly to each candidate, with gray denoting an explored endpoint, yellow an unexplored endpoint, and cyan an incompletely explored junction.

\subsection{Prompt and evaluation protocol}
We adopt the base scoring prompt from WMNav~\citep{wmnav}, as it provides an existing VLM-based navigation prompt for evaluating candidate directions from observations and does not encode exploration memory itself. This gives a common reference prompt across all representations and avoids tailoring the base instruction to our proposed memory interface. All variants use the same scoring instruction and differ only in the minimal representation-specific information required by each interface. The complete prompts are provided in Fig.\ref{prompt_toy}. The VLM scores each displayed candidate from $0$ to $10$, with higher scores indicating a stronger preference for further exploration. \textit{Textual History} adds the history summary, while \textit{Visited Markers} and \textit{Choice-Aligned State} add only the corresponding marker definitions. We evaluate Gemini-3.5-Flash and Qwen3.5-Plus with temperature $0$ and thinking disabled. Each representation is queried independently $10$ times per model. Repeated calls can still produce different outputs, particularly for Qwen3.5-Plus, so Fig.~\ref{motivation} reports the mean and standard deviation of the candidate scores across runs. These scores are used only to make changes in VLM preference visible in this motivation study and are not part of the MarvisNav navigation pipeline. The complete raw responses will be provided along with the code.

\section{Navigation and Ablation Prompts}
\label{app:navigation_prompts}

\begin{figure}[h]
    \centering
    \includegraphics[width=0.95\linewidth]{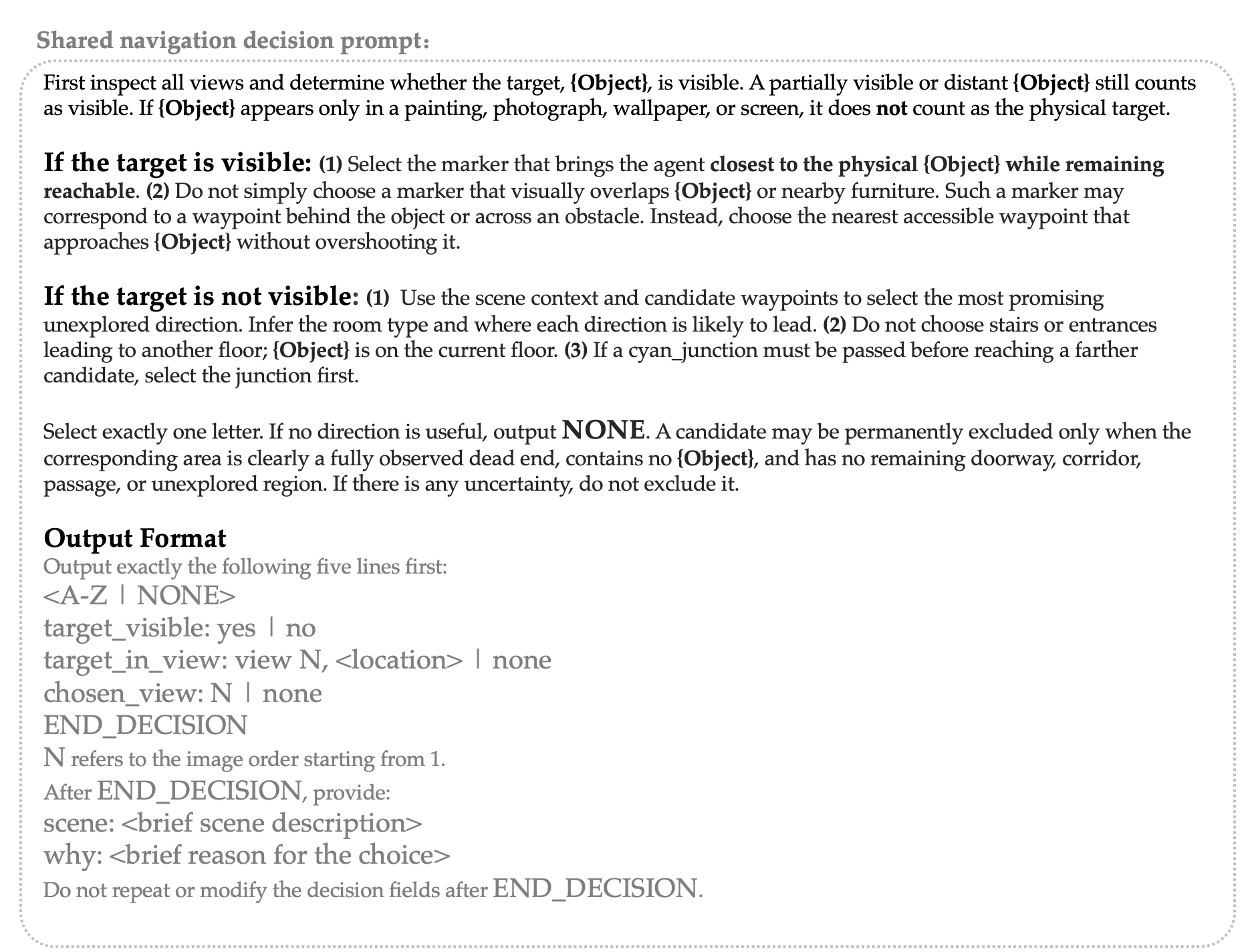}
    \caption{Shared navigation decision prompt}
    \label{shared}
\end{figure}

For the full prompt used in navigation experiments, we replace the shared scoring block
in Fig.~\ref{prompt_toy} with the navigation decision
prompt shown in Fig.~\ref{shared}.
This prompt instructs the VLM to select a route candidate and report target visibility.
The memory-representation ablations follow the same modular prompt structure as Appendix~\ref{motivation_details}, with memory descriptions
and marker definitions specific to each representation.

\section{Failure Case Analysis}

\begin{figure*}[h]
    \centering
    \begin{subfigure}[t]{0.33\textwidth}
        \centering
        \includegraphics[width=\linewidth]{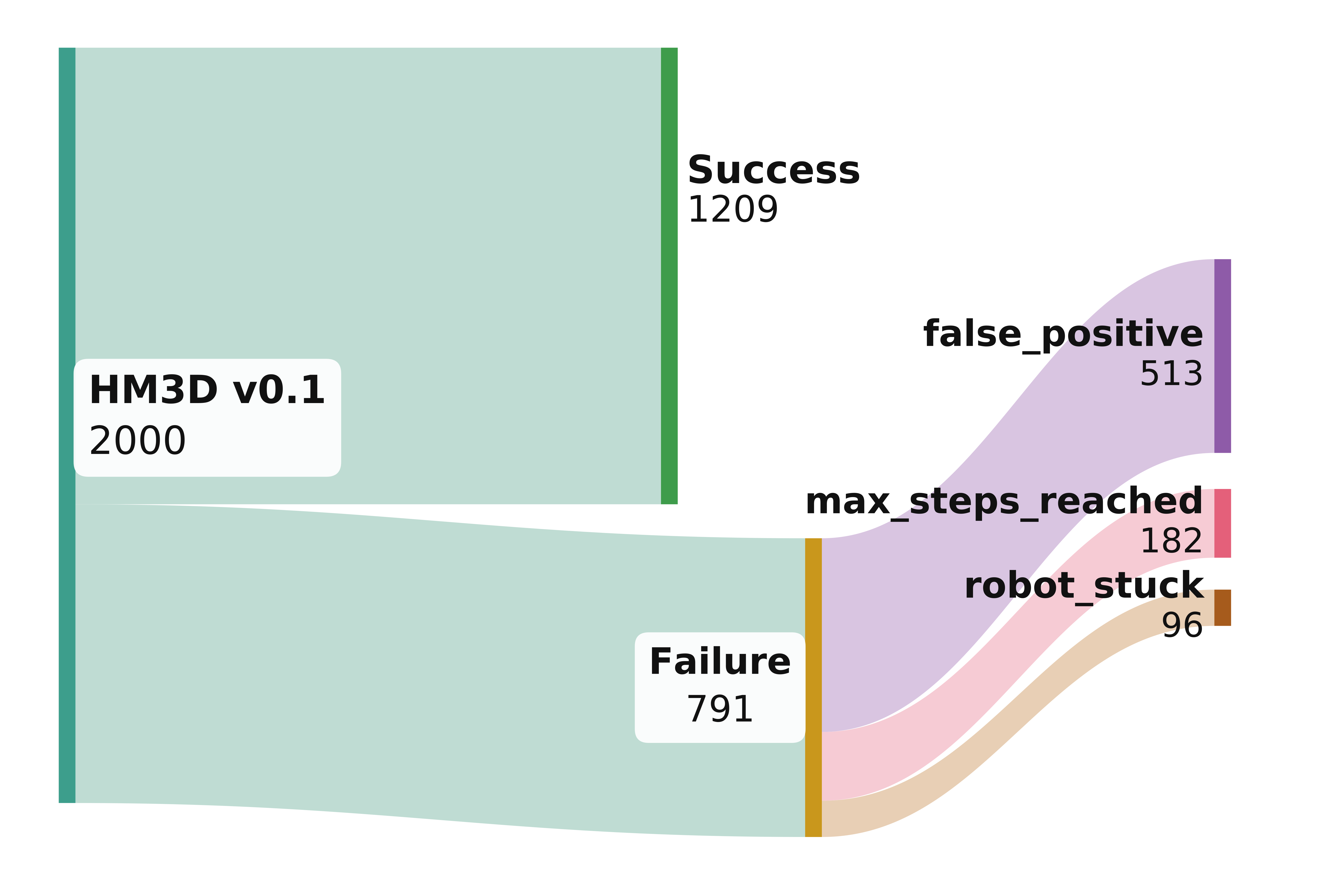}
        \label{fig:hm3d_original_f}
    \end{subfigure}
    \hfill
    \begin{subfigure}[t]{0.33\textwidth}
        \centering
        \includegraphics[width=\linewidth]{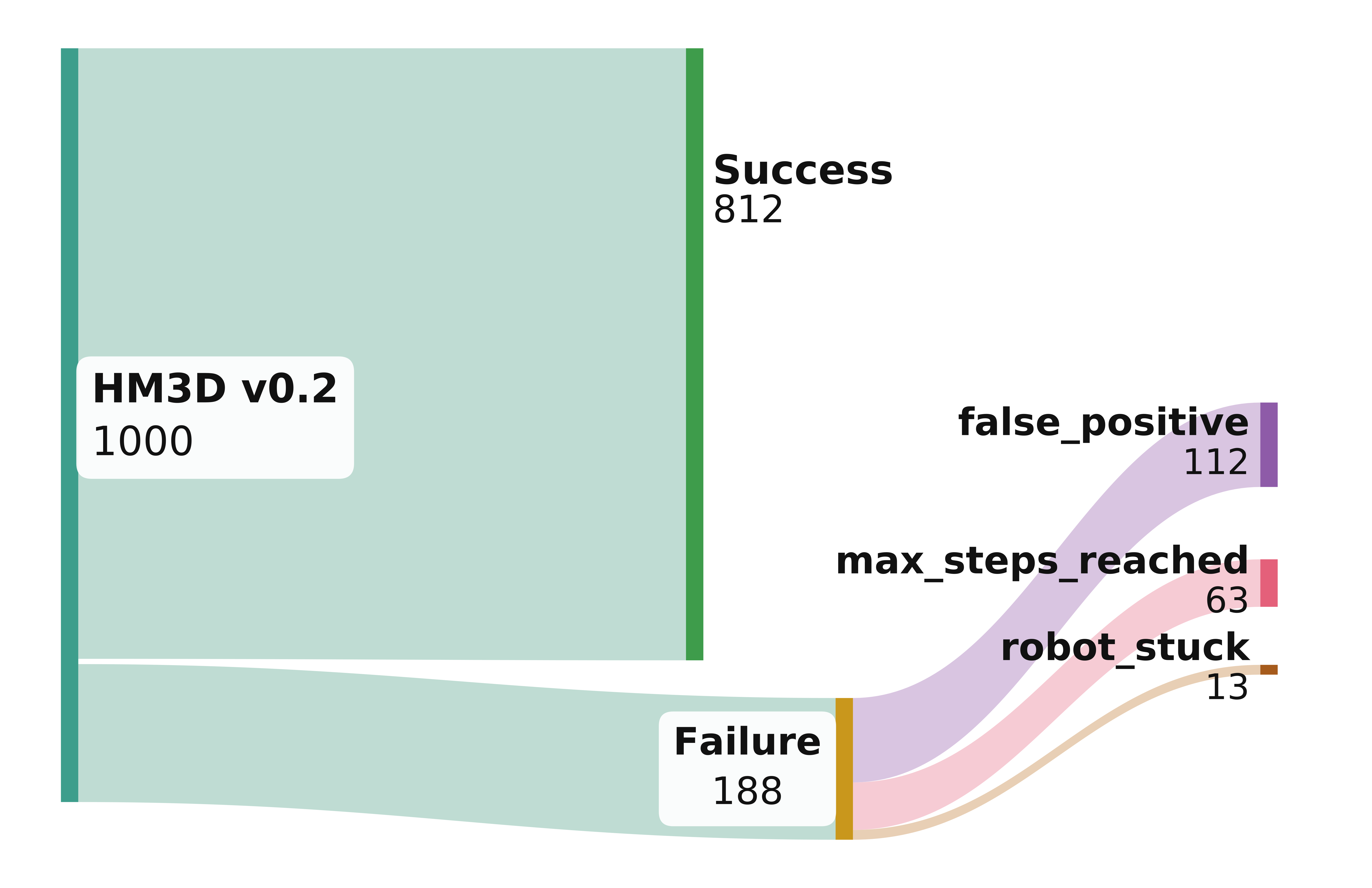}
        \label{fig:hm3d_reconstruction_f}
    \end{subfigure}
    \hfill
    \begin{subfigure}[t]{0.3\textwidth}
        \centering
        \includegraphics[width=\linewidth]{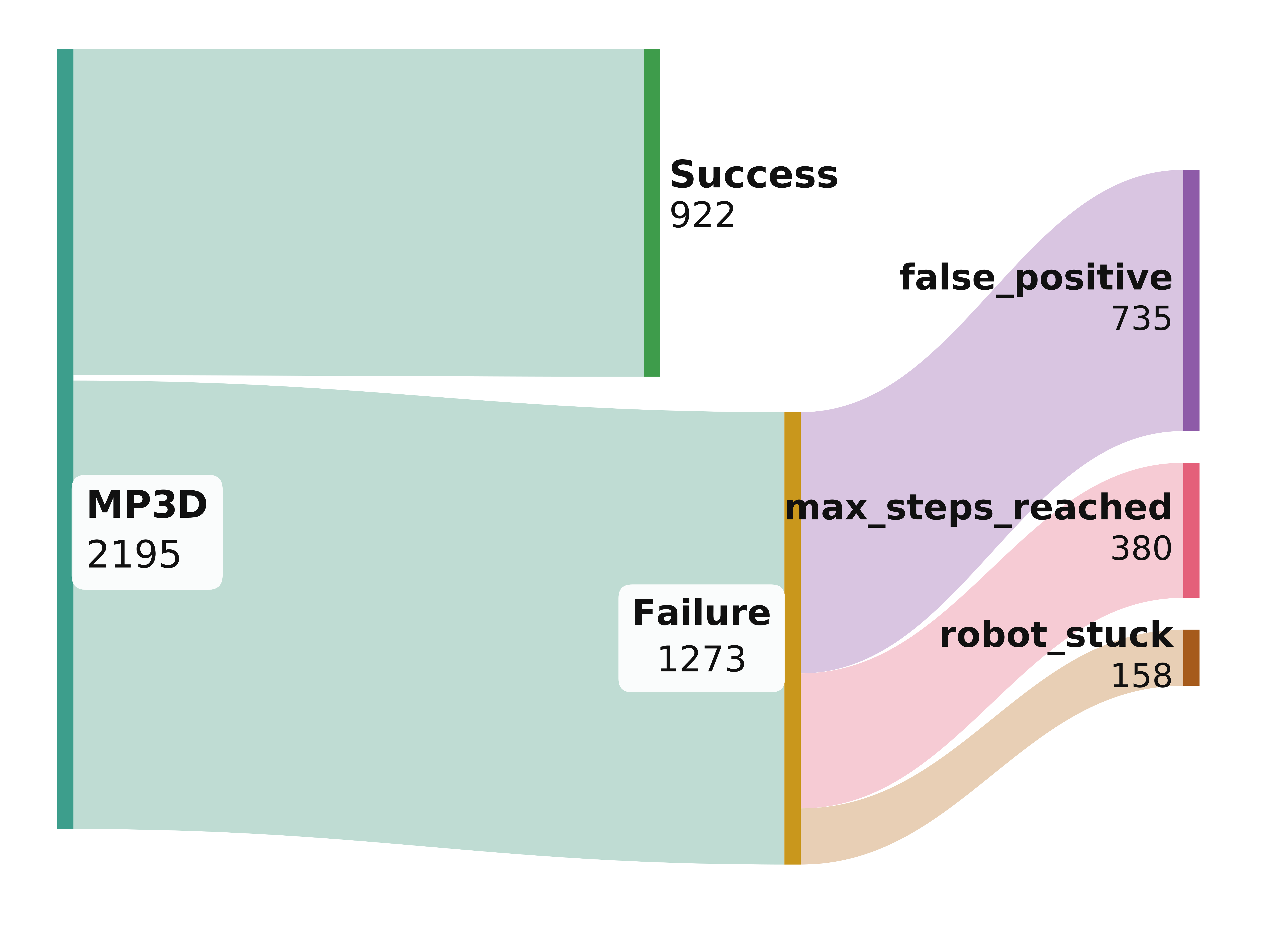}
        \label{fig:hm3d_rendered_f}
    \end{subfigure}

    \caption{\textbf{Failure analysis across HM3D v0.1, HM3D v0.2, and MP3D.}
Evaluation episodes are decomposed into successes and three failure outcomes: false-positive target detection, maximum step budget reached, and robot stuck. Numbers denote episode counts.}
    \label{failure}
\end{figure*}

To better understand the remaining limitations of MarvisNav, we analyze unsuccessful episodes on HM3D v0.1, HM3D v0.2, and MP3D using the recorded termination reasons as shown in Fig.\ref{failure}. Despite different overall success rates, the failure composition is broadly consistent across benchmarks. False-positive target detections are the dominant failure mode, accounting for $64.9\%$, $59.6\%$, and $57.7\%$ of failures on HM3D v0.1, HM3D v0.2, and MP3D, respectively. Step-budget exhaustion is the second most common outcome ($23.0\%$--$33.5\%$), while robot-stuck failures remain relatively infrequent ($6.9\%$--$12.4\%$). The prevalence of false positives indicates that target perception remains an important bottleneck beyond high-level route selection. Step-budget exhaustion can result from inefficient exploration, including missed target detections, or simply from the scale of the environment. The latter is particularly relevant on MP3D, where increasing the maximum step budget may alleviate some failures. Robot-stuck cases are largely caused by imperfections in the scanned benchmark geometry, such as holes or discontinuities that disrupt low-level navigation. Overall, these results suggest that improving target perception and recovery from long-horizon failures are promising directions for further improvement in future work.

\begin{table*}[h]
\caption{\textbf{Benchmark issues identified in the HM3D v0.2 evaluation split.}
We report the $52$ identified episodes grouped by issue type and scene.}
\label{tab:hm3d_problematic_episodes}
\centering
\small
\setlength{\tabcolsep}{3pt}
\renewcommand{\arraystretch}{0.95}
\begin{tabular}{@{}p{3.2cm}lp{4.5cm}c@{}}
\toprule
\textbf{Issue Type} &
\textbf{Scene ID} &
\textbf{Episode IDs} &
\textbf{Number of Episodes} \\
\midrule

No valid target in reachable region
& 7MXmsvcQjpJ
& 84, 89, 96, 109
& 4 \\

\midrule

\multirow[t]{14}{3.2cm}{Annotation error / omission}
& Dd4bFSTQ8gi
& 196, 197, 222
& 3 \\
& bCPU9suPUw9
& 508, 511, 528
& 3 \\
& jxsvVRusffK
& 538, 539, 558
& 3 \\
& p53SfW6mjZe
& 732
& 1 \\
& a8BtkwhxdRV
& 492, 499
& 2 \\
& GLAQ4DNUx5U
& 228, 238, 243
& 3 \\
& EF36g7L6Z9M
& 595, 610, 611
& 3 \\
& mL8ThkuaVTM
& 673
& 1 \\
& svBbv1Pavdk
& 841, 843, 846, 863
& 4 \\
& BAbdmeyTvMZ
& 119
& 1 \\
& cvZr5TUy5C5
& 566
& 1 \\
& 5cdEh9F2hJL
& 32, 44
& 2 \\
& HY1NcmCgn3n
& 261, 262, 263, 264, 265, 266, 267, 268, 269
& 9 \\
& yr17PDCnDDW
& 946, 948
& 2 \\

\midrule

\multirow[t]{3}{3.2cm}{Incomplete scene reconstruction}
& HY1NcmCgn3n
& 254, 255, 259, 260
& 4 \\
& bCPU9suPUw9
& 520, 527
& 2 \\
& k1cupFYWXJ6
& 644, 646, 648, 649
& 4 \\

\midrule
\multicolumn{3}{l}{\textbf{Total problematic episodes}} & \textbf{52} \\
\bottomrule
\end{tabular}
\end{table*}

\begin{figure*}[h]
    \centering
    \begin{subfigure}[t]{1\textwidth}
        \centering
        \includegraphics[width=\linewidth]{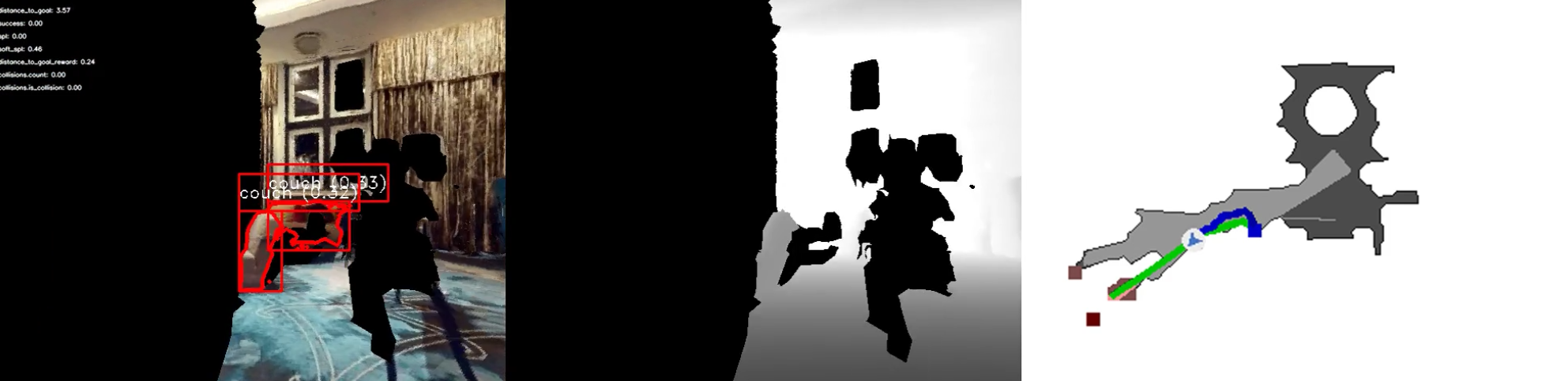}

        \caption{\textbf{k1cupFYWXJ6}, ep 649}
    \end{subfigure}
    \hfill
    \begin{subfigure}[t]{1\textwidth}
        \centering
        \includegraphics[width=\linewidth]{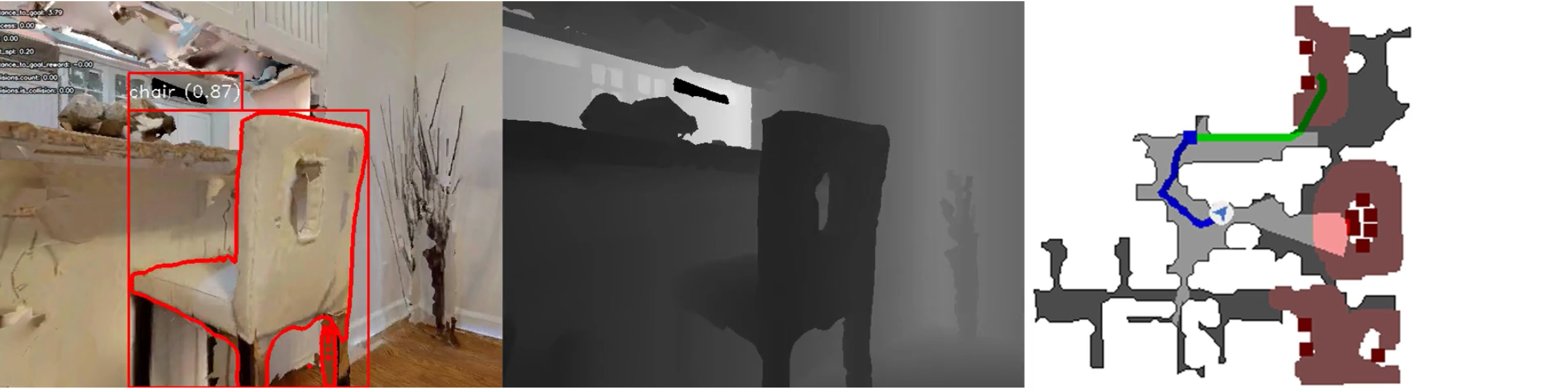}

        \caption{\textbf{Dd4bFSTQ8gi}, ep 197}
    \end{subfigure}
    \hfill
    \begin{subfigure}[t]{1\textwidth}
        \centering
        \includegraphics[width=\linewidth]{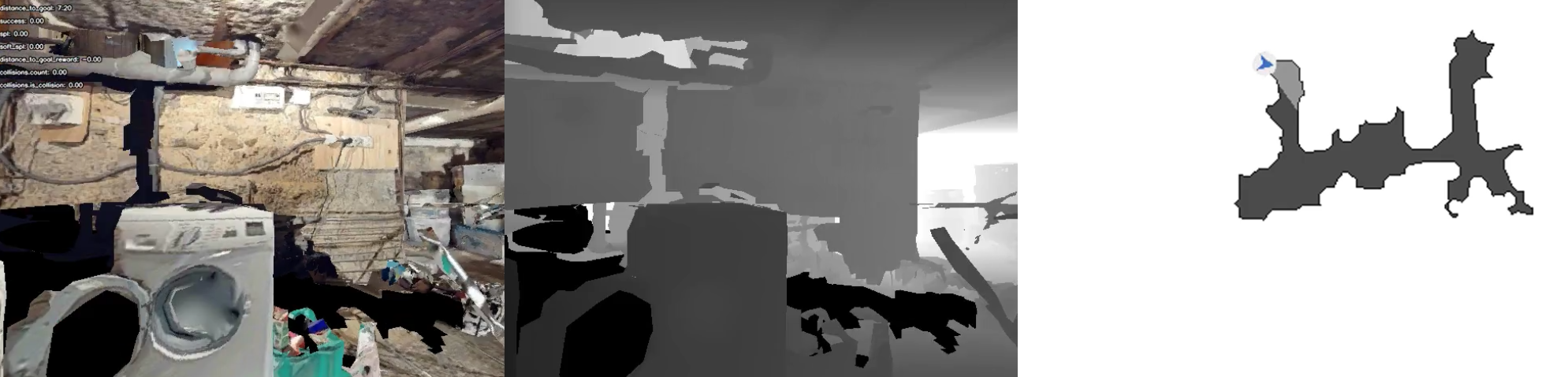}

        \caption{\textbf{7MXmsvcQjpJ}, ep 96}    
    \end{subfigure}

\caption{\textbf{Representative benchmark-related failure cases from HM3D v0.2.}
Each row shows a representative problematic episode, including the egocentric observation, the corresponding depth/geometry view, and a top-down scene map. In the third column, red regions indicate the benchmark ground-truth object region.
\textbf{(a) Incomplete scene reconstruction} (\texttt{k1cupFYWXJ6}, ep 649, target: \textit{sofa}): missing or corrupted scene geometry around the target region makes the episode unreliable for evaluation.
\textbf{(b) Annotation error / omission} (\texttt{Dd4bFSTQ8gi}, ep 197, target: \textit{chair}): the agent successfully reaches a visible chair, but the benchmark ground-truth annotation is incomplete or inconsistent, so the episode is incorrectly counted as a false positive.
\textbf{(c) No valid target in the reachable region} (\texttt{7MXmsvcQjpJ}, ep 96, target: \textit{sofa}): the episode asks the agent to search for a sofa in a basement area where no valid target instance exists. These examples illustrate three representative benchmark defects that can cause apparent navigation failures unrelated to policy quality.}
    \label{failure_case}
\end{figure*}

As HM3D v0.2 has become an common evaluation benchmark for ZSON methods~\citep{apexnav,openfrontier}, the quality of its evaluation episodes directly affects cross-method comparison. To support more reliable evaluation, we manually inspect all $1{,}000$ episodes in the HM3D v0.2 evaluation split and identify $52$ episodes with benchmark-related issues. These cases fall into three main categories: incomplete scene reconstruction, including large holes, missing geometry, severely degraded small objects, or geometry that makes targets unreachable; inconsistent target annotations, where the annotated category does not match the visible object; and episodes in which no valid target instance exists in the agent's reachable region. The representative cases are shown in Fig \ref{failure_case}. Such defects can turn benchmark limitations into apparent navigation failures, and these $52$ episodes therefore impose an evaluation ceiling unrelated to navigation policy quality. We provide the corresponding scene and episode IDs together in Table~\ref{tab:hm3d_problematic_episodes} with their issue categories to support independent verification, reproducible comparison, and future benchmark curation by the community, and recommend excluding the verified invalid episodes when constructing a cleaned HM3D v0.2 evaluation split. More broadly, we hope this analysis supports more reliable ObjectNav evaluation by helping separate algorithmic failures from artifacts of benchmark construction. It also highlights the need for better-curated and more reliable benchmarks in future research.

\section{Future Work}

Future work could extend the local exploration summaries in the State-Aware RVG to hierarchical summaries that combine node- and region-level progress while keeping memory aligned with current visual choices. In this work, we use RVG to construct memory-bearing visual route choices, but the same design could be adapted to other topological graphs. Future work could also extend the spatial representation to support multi-floor object search \citep{werby23hovsg}. Another direction is to improve geometric reliability around glass and reflective surfaces through uncertainty-aware depth estimation and multi-view verification~\citep{zheng2026enhancing}, providing more reliable geometry for RVG construction and the projection of visual route candidates into egocentric views. Finally, MarvisNav's candidate-selection interface motivates exploring lower-latency decision backends, including real-time multimodal interaction through the Gemini Live and visual extensions inspired by Jev's structured-decision design~\cite{almeida2026jev}. These advances may enable VLM-based ZSON to approach the responsiveness of end-to-end navigation policies without task-specific policy training.

\end{document}